\documentclass[11pt,a4paper]{article}
\usepackage[hyperref]{acl2021}
\usepackage{times}
\usepackage{latexsym}

\usepackage{graphicx}
\usepackage{amsmath}
\usepackage{mathrsfs}
\usepackage{amsfonts}
\usepackage{multirow}
\usepackage{booktabs}
\usepackage{subcaption}
\usepackage{lipsum}
\usepackage{booktabs,subcaption,amsfonts,dcolumn}
\newcolumntype{d}[1]{D..{#1}}

\usepackage{algorithm}  
\usepackage{algpseudocode}  
\usepackage{amsmath}  
\usepackage{color}

\usepackage{microtype}

\aclfinalcopy 
\def\aclpaperid{1778} 

\title{UNIMO: Towards Unified-Modal Understanding and Generation via Cross-Modal Contrastive Learning}

\author{Wei Li\thanks{\ \ These authors contribute equally to this study and are listed with random order.}, Can Gao\footnotemark[1], Guocheng Niu\footnotemark[1],  Xinyan Xiao\footnotemark[1], \\ \textbf{Hao Liu, Jiachen Liu, Hua Wu, Haifeng Wang} \\
  Baidu Inc., Beijing, China \\
  \texttt{\{liwei85,gaocan01,niuguocheng,xiaoxinyan,} \\
  \texttt{liuhao24,liujiachen,wu\_hua,wanghaifeng\}@baidu.com}
  }

\date{}

\begin{document}
\maketitle
\begin{abstract}
Existed pre-training methods either focus on single-modal tasks or multi-modal tasks, and cannot effectively adapt to each other.
They can only utilize single-modal data (i.e., text or image) or limited multi-modal data (i.e., image-text pairs).
In this work, we propose a UNIfied-MOdal pre-training architecture, namely UNIMO, which can effectively adapt to both single-modal and multi-modal understanding and generation tasks.
Large scale of free text corpus and image collections are utilized to improve the capability of visual and textual understanding, and cross-modal contrastive learning (CMCL) is leveraged to align the textual and visual information into a unified semantic space, over a corpus of image-text pairs augmented with related images and texts.
With the help of rich non-paired single-modal data, our model is able to learn more generalizable representations, by allowing textual knowledge and visual knowledge to enhance each other in the unified semantic space.
The experimental results show that UNIMO greatly improves the performance of several single-modal and multi-modal downstream tasks.
Our code and pre-trained models are public at the UNIMO project page \url{https://unimo-ptm.github.io/}.

\end{abstract}

\section{Introduction}
\label{sec:intro}

Large-scale pre-training has drawn much attention in both the community of Compute Vision (CV) and Natural Language Processing (NLP) due to its strong capability of generalization and efficient usage of large-scale data.
Firstly in CV, a series of models were designed and pre-trained on the large-scale dataset ImageNet, such as AlexNet \citep{krizhevsky2017imagenet}, VGG \citep{simonyan2014very} and ResNet \citep{he2016deep}, which effectively improved the capability of image recognition for numerous tasks.
Recent years have witnessed the burst of pre-training in NLP, such as BERT \citep{devlin-etal-2019-bert}, RoBERTa \citep{liu2019roberta}, XLNet \citep{yang2019xlnet} and UniLM \citep{dong2019unified}, which greatly improve the capabilities of language understanding and generation.
However, the above researches focus on the single-modal learning and can only be effectively used in single-modal (i.e., only text or image) scenarios.
In order to adapt to multi-modal scenarios, a series of multi-modal pre-training methods were proposed and pre-trained on the corpus of image-text pairs, such as ViLBERT \citep{lu2019vilbert}, VisualBERT \citep{li2019visualbert} and UNITER \citep{chen2020uniter}, which greatly improve the ability to process multi-modal information.
However, these models can only utilize the limited corpus of image-text pairs and cannot be effectively adapted to single-modal scenarios \citep{lin2020interbert}.

\begin{figure}[t!]
	\centering
	\includegraphics[width=3in]{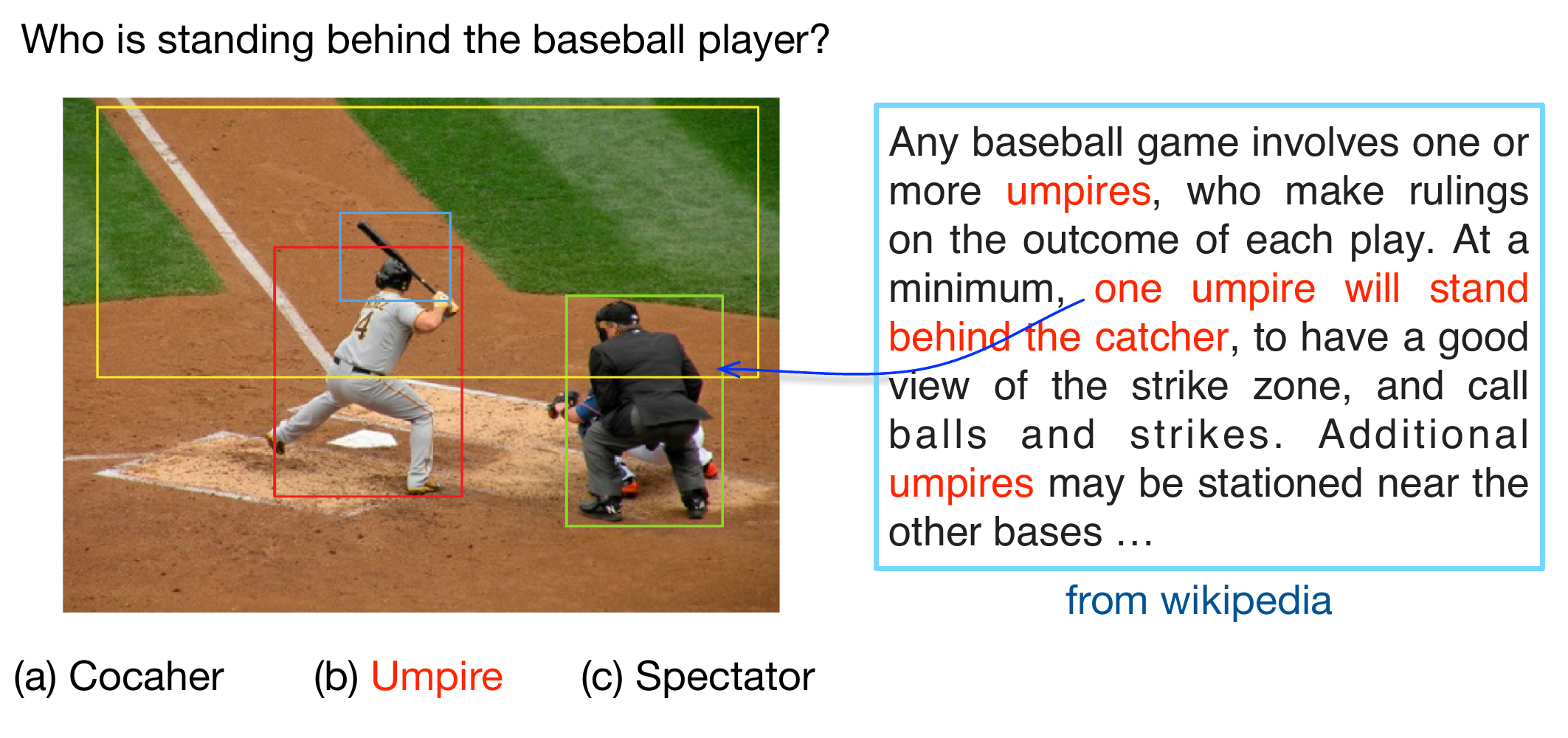}
	\caption{An illustrative example for the necessity of unified-modal learning. We can only determine the correct answer to the visual question based on the textual background information.}
	\label{fig:example}
\end{figure}

\begin{figure}[t!]
	\centering
	\includegraphics[width=3.1in]{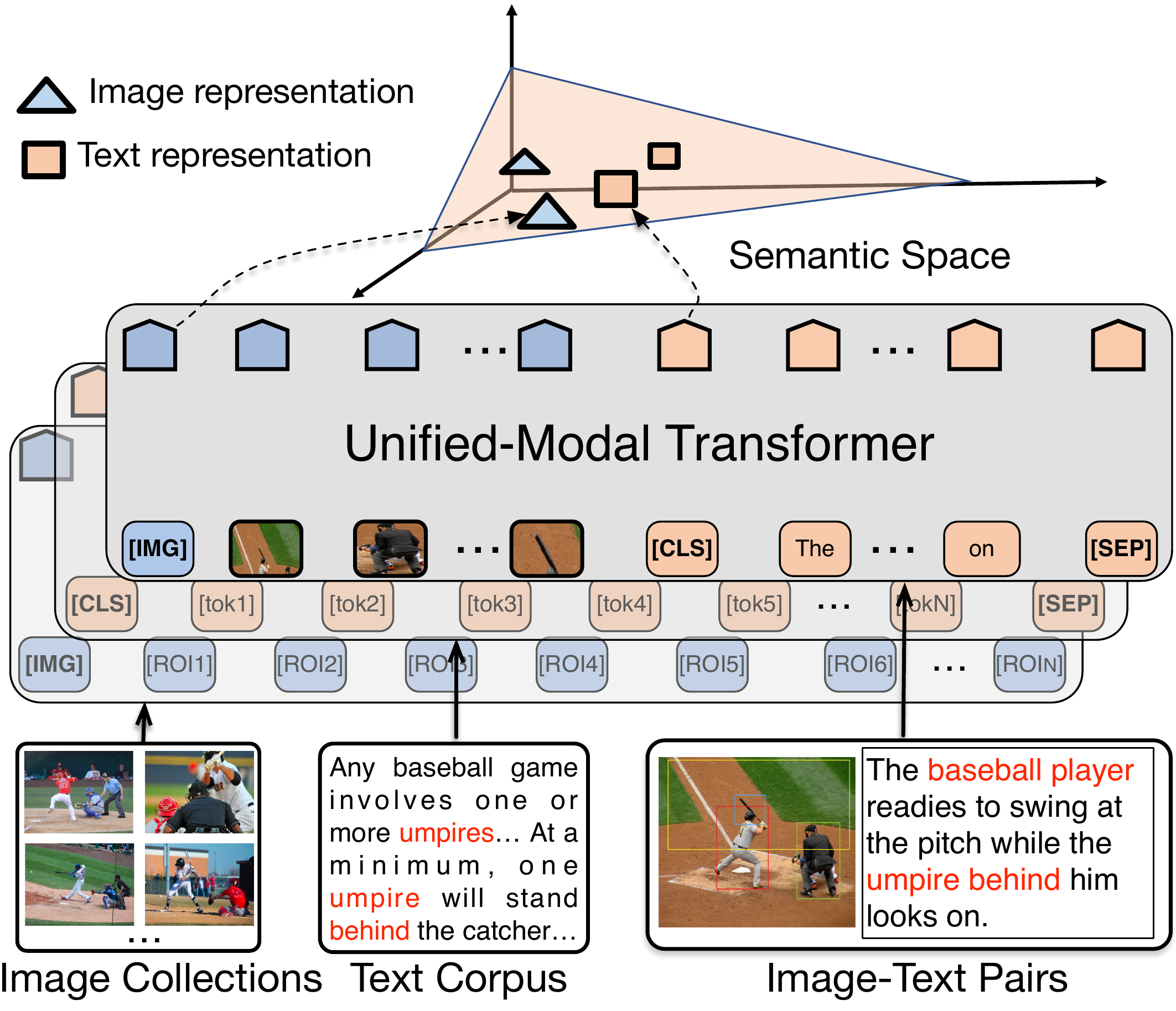}
	\caption{Illustration of the unified-modal pre-training architecture. Both image collections, text corpus and image-text pairs can be effectively utilized for representation learning.}
	\label{fig:unifmodal}
\end{figure}

A smarter AI system should be able to process different modalities of information effectively.
There are large scale of data in different modalities on the Web, mainly textual and visual information.
The textual knowledge and the visual knowledge usually can enhance and complement each other.
As the example shown in Figure \ref{fig:example}, it's difficult to answer the question correctly only with the visual information in the image. 
However, if we connect the visual information to the textual information which describes the background of a baseball game, it's very easy to determine the correct answer.
Also, the visual information can make it easier to understand the scene described by the text.
The research in neuroscience by \citet{van2018neuronal} reveals that the parts of the human brain responsible for vision can learn to process other kinds of information, including touch and sound.
Inspired by this research, we propose to design a unified-modal architecture UNIMO which aims to process multi-scene and multi-modal data input with one model, including textual, visual and vision-and-language data, as shown in Figure \ref{fig:unifmodal}.

The greatest challenge to unify different modalities is to align and unify them into the same semantic space which are generalizable to different modalities of data.
Existed cross-modal pre-training methods try to learn cross-modal representations based on only limited image-text pairs by simple image-text matching and masked language modeling \citep{chen2020uniter}.
They can only learn specific representations for image-text pairs, and thus fail to generalize to single-modal scenarios.
So their performance will drop dramatically when applied to language tasks \citep{lin2020interbert}, which has also been revealed by our experiments (see Section \ref{ssec:sin}).
In this work, UNIMO learns visual representations and textual representations simultaneously, and unifies them into the same semantic space via cross-modal contrastive learning (CMCL) based on a large-scale corpus of image collections, text corpus and image-text pairs.

\begin{figure*}[t!]
	\centering
	\includegraphics[width=6in]{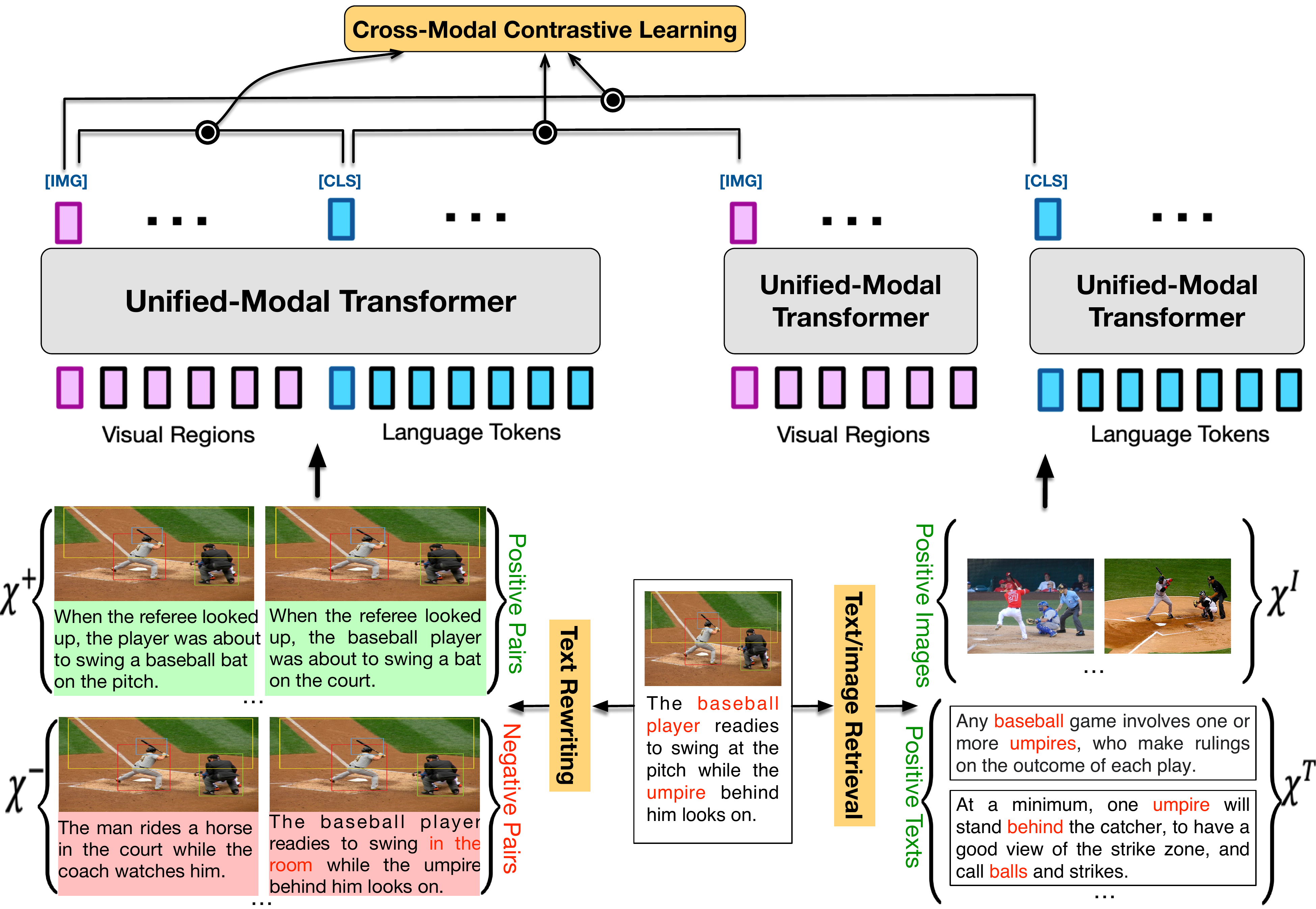}
	\caption{Illustration of the CMCL. A series of text rewriting techniques are utilized to create positive image-text pairs $\mathcal{X^+}$ and hard negative image-text pairs $\mathcal{X^-}$. Image and text retrieval are also utilized to obtain related images $\mathcal{X^I}$ and texts $\mathcal{X^T}$ from single-modal data, which are treated as single-modal positive samples during cross-modal learning. All of them are encoded by the same unified-modal Transformer in pairs or individually, and the representations of images and texts are extracted to compute the contrastive loss.}
	\label{fig:cmcl-sg}
\end{figure*}

UNIMO effectively utilizes the large-scale of text corpus and image collections to learn general textual and visual representations. 
The CMCL aligns the visual representations and textual representations, and unifies them into the same semantic space based on image-text pairs.
As shown in Figure \ref{fig:cmcl-sg}, to facilitate different levels of semantic alignment between vision and language, we propose to utilize a series of text rewriting techniques to improve the diversity of cross-modal information.
Specifically, for an image-text pair, various positive examples and hard negative examples can be obtained by rewriting the original caption at different levels.
Moreover, to incorporate more background information from the single-modal data, text and image retrieval are also applied to augment each image-text pair with various related texts and images.
The positive pairs, negative pairs, related images and texts are learned jointly by CMCL.
In this way, our model can effectively unify different levels of visual and textual representations into the same semantic space, and incorporate more single-modal knowledge to enhance each other.

The unified-modal architecture mainly has the following advantages compared with previous methods:
\begin{itemize}
\item We can utilize large scale of non-paired text corpus and image collections on the Web to learn more generalizable textual and visual representations, and improve the capability of vision and language understanding and generation.
\item Our model can be effectively fine-tuned for both single-modal and multi-modal understanding and generation downstream tasks.
\item The visual knowledge and textual knowledge can enhance each other to achieve better performance on several single-modal and multi-modal tasks than previous methods.
\end{itemize}

\section{UNIMO}
\label{sec:unimo}

Humans perceive the world through many modalities, such as sound, vision and language. 
Even though any individual modality might be incomplete or noisy, important information is still perceivable since they tend to be shared or enhanced each other. 
With this motivation, we propose a unified-modal pre-training method UNIMO to learn representations that capture modality-invariant information at the semantic level.
Different from previous methods, UNIMO learns from different modalities of data, including images, texts and image-text pairs, thus achieving more robust and generalizable representations for both textual and visual input.

As shown in Figure \ref{fig:unifmodal}, UNIMO employs multi-layer self-attention Transformers to learn unified semantic representations for both textual and visual data.
For a textual input W, it is firstly split into a sequence of subwords $W=\{[CLS],w_1,...,w_n,[SEP]\}$ by Byte-Pair Encoding (BPE) \citep{sennrich-etal-2016-neural}, and then the self-attention mechanism is leveraged to learn contextual token representations $\{h_{[CLS]},h_{w_1}, ...,h_{w_n},h_{[SEP]}\}$. The special tokens $[CLS]$ and $[SEP]$ denote the start and end of the textual sequence, respectively.
Similarly, for an image V, it is firstly converted to a sequence of region features $V=\{[IMG],v_1, ...,v_t\}$ ($[IMG]$ denotes the representation of the entire image), and then the self-attention mechanism is leveraged to learn contextual region representations $\{h_{[IMG]}, h_{v_1}, ..., h_{v_t}\}$.
Similar to previous work \citep{chen2020uniter}, we use Faster R-CNN \citep{ren2016faster} to detect the salient image regions and extract the visual features (pooled ROI features) for each region.
For an image-text pair $(V, W)$, its visual features and textual tokens are concatenated as a sequence $\{[IMG], v_1, ..., v_t, [CLS],w_1,...,w_n,[SEP]\}$.
Then the sequence is feed into the multi-layer Transformer network to learn cross-modal contextual representations for both the textual tokens and image regions. 
We extract the representations $h_{[IMG]}$ and $h_{[CLS]}$ as the semantic representations of image $V$ and text $W$, respectively.

Based on large volumes of image collections $\{V\}$, text corpus $\{W\}$ and image-text pairs $\{(V,W)\}$, UNIMO learns generalizable visual and textual representations in similar ways by masked prediction, and unify them into the same semantic space via CMCL.
Joint visual learning on image collections, language learning on text corpus and cross-modal learning on image-text pairs not only improve the capability of visual and language understanding and generation, but also enable the textual knowledge and visual knowledge to enhance each other in the unified semantic space.

\subsection{Cross-Modal Contrastive Learning}
\label{ssec:cmcl}

The greatest challenge to unify different modalities is to align and unify their representations at different levels.
For the example shown in Figure \ref{fig:unifmodal}, the model not only needs to connect the scene shown in the whole image to an article describing a baseball game, but also needs to align the two men and their location relationship in the image with ``baseball player'', ``umpire'' and ``behind'' in the text, respectively.
Several existing cross-modal pre-training methods try to align visual and textual representations by simply image-text matching \citep{li2019unicoder, chen2020uniter} based on a limited corpus of image-text pairs.
They randomly sample a negative image or text from the same training batch for each image-text pair, and utilize a classifier to determine whether the image and text are matching.
As the randomly sampled negative text or image is usually very different from the original text or image, they can only learn very coarse alignment between textual and visual representations.
In this work, we propose a novel CMCL method to align and unify different levels of textual and visual representations into the same semantic space.

The main idea is to let the representations of the paired image and text near in the representation space while the non-paired far away.
The representations of image $V$ and text $W$ are used to compute the similarity between them to measure their distance $d(V, W)$.
As shown in Figure \ref{fig:cmcl-sg}, to facilitate semantic alignment between vision and language at different levels, we design several novel text rewriting techniques to rewrite the original caption of an image either at word, phrase or sentence level.
In this way, we can create large volumes of positive examples $\mathcal{X^+}$ and negative examples $\mathcal{X^-}$ for each image-text pair $(V, W)$.
Moreover, to augment cross-modal learning with single-modal information, text and image retrieval are applied to obtain various related texts $\mathcal{X}^T$ and images $\mathcal{X}^I$ for each image-text pair $(V, W)$.
Different from the positive and negative image-text pairs, the retrieved images and texts are encoded individually as they  mainly carry weak correlations, as shown in the right part of Figure \ref{fig:cmcl-sg}.
Based on these positive and negative examples, the following contrastive loss $\mathcal{L}_{CMCL}$ is utilized to learn detailed semantic alignments across vision and language:
\begin{equation}
\small
\begin{aligned}
      \mathbb{E}_{V,W} \left[ -log \frac{\sum_{(V^+,W^+) \in \mathcal{X}^{\{+,I,T\}}} exp(d(V^+,W^+)/ \tau)} {\sum_{(V',W') \in \mathcal{X}^{\{-,+,I,T\}}} exp(d(V',W')/ \tau)} \right]
\end{aligned}
\label{eq-cmcl}
\end{equation}
where $\tau$ denotes the temperature parameter. Note that, for single-modal images $\mathcal{X}^I$ and texts $\mathcal{X}^T$, the original text $W$ and image $V$ are used to compute the cross-modal relevance, respectively. To the best of our knowledge, this is the first work that explores CMCL to unify visual and textual semantic space.

\paragraph{Text Rewriting}
To enhance multi-granularity of semantic alignment between image and text, we rewrite the caption of an image at different levels, including sentence-level, phrase-level and word-level.
For sentence-level rewriting, we utilize the back-translation techniques \citep{edunov-etal-2018-understanding} to obtain several positive samples for each image-text pair.
Specifically, each caption of an image is translated into another language and then translated back to the original language.
In this way, several similar captions can be obtained for an image.
Furthermore, for each image-text pair, the most similar captions of other images are retrieved based on TF-IDF similarity.
The retrieved results are very similar to the original caption but doesn't accurately describe the corresponding image, so they can be used as hard negative samples to enhance the sentence-level alignment between image and text.
For phrase-level and word-level rewriting, we first parse the image caption into a scene graph \citep{wang-etal-2018-scene}, then randomly replacing the object, attribute or relation nodes of the scene graph with a different object, attribute or relation from the corresponding vocabularies.
Instead of randomly sampling negative samples as previous methods, text rewriting can generate large volumes of hard negative samples.
In this way, we can help the model to learn more detailed semantic alignment from different levels between image and text.

\paragraph{Image/Text Retrieval}
In order to incorporate more single-modal information during cross-modal learning, each image-text pair is further augmented with various related images and texts that retrieved from the single-modal data. 
Specifically, for an image, other images in the image collections will be ordered by their visual similarities.
Those images that have highly overlapped objects with the original image will be extracted to provide relevant visual information.
Similarly, sentences that are semantically related with the original caption are extracted based on semantic similarity to provide background language information.
The retrieved images and texts are encoded individually by the unified-modal Transformer as shown in Figure \ref{fig:cmcl-sg}, then their representations are extracted to compute the cross-modal contrastive loss in Equation \ref{eq-cmcl}.
These retrieved single-modal information provide rich background information for better cross-modal learning.

\subsection{Visual Learning}
\label{ssec:vis}

Similar to the masked language modeling in BERT, we sample image regions and mask their visual features with a probability of $15\%$.
The visual features of the masked regions are replaced by zeros.
As the regions from an image usually are highly overlapped with each other, we choose to mask all regions that have a high proportion of mutual intersection to avoid information leakage.
Similar to \citet{lin2020interbert}, we randomly choose regions as masking anchors and mask the regions whose overlapping ratios with the anchors are larger than 0.3.
For an image $V$, the model is trained to reconstruct the masked regions $v_m$ given the remaining regions $v_{\backslash m}$:
\begin{equation}
\begin{aligned}
    \mathcal{L}_{V} = \mathbb{E}_{V \in D} f_{\theta} (v_m|v_{\backslash m})
\end{aligned}
\label{eq1}
\end{equation}
Similarly, for an image-text pair $(V,W)$, the model is trained to reconstruct the masked regions $v_m$ given the text $W$ and the remaining regions $v_{\backslash m}$:
\begin{equation}
\begin{aligned}
    \mathcal{L}_{V} = \mathbb{E}_{V,W \in D} f_{\theta} (v_m|v_{\backslash m}, W)
\end{aligned}
\label{eq1}
\end{equation}

As the visual features are high-dimensional and continuous, we utilize both feature regression and region classification objective to learn better visual representations.
The feature regression learns to regress the contextualized visual representations $h_{v_i}$ to its visual features $v_i$, which can be formulated as: $f_{\theta} (v_m|v_{\backslash m}) = \sum_{i=1}^{M} \Vert r(h_{v_i}) - v_i \Vert^2$, where $r$ indicates an FC layer to convert $h_{v_i}$ into a vector of the same dimension as $v_i$.
The region classification learns to recognize the object semantic class of each masked region based on its contextualized visual representation $h_{v_i}$.
An FC layer is utilized to compute the scores for K object classes $s(h_{v_i})$, which further goes through a $softmax$ function to obtain the normalized distribution.
The final objective minimizes the cross-entropy (CE) loss between the predicted distribution and the object detection output $c(v_i)$ from Faster R-CNN: $f_{\theta} (v_m|v_{\backslash m}) = \sum_{i=1}^{M} CE(softmax(s(h_{v_i})), c(v_i))$.
The score function $f_{\theta} (v_m|v_{\backslash m},W)$ is formulated similarly.

\subsection{Language Learning}
\label{ssec:lan}

To learn general language representations for both language understanding and generation tasks, our model is trained as a unified encoder-decoder model with two types of language modeling tasks: bidirectional prediction and sequence-to-sequence (Seq2Seq) generation.
The unified modeling is achieved by utilizing specific self-attention masks to control what context the prediction conditions on, inspired by \citet{dong2019unified}.
To improve the language learning process, we firstly detect semanticly complete phrases from the text, such as name entities by syntactic parsing, and then treat them as a whole in the following masking strategies.
Different from previous work, we always sample a sequence of complete words or phrases instead of subword tokens, for both bidirectional prediction and Seq2Seq generation.

\paragraph{Bidirectional prediction.} 
Given a sequence of tokens $W=\{[CLS],w_1,...,w_n,[SEP]\}$, we iteratively sampling spans of text until totally $15\%$ tokens have been selected.
We sample the span length from a geometric distribution $l \sim Geo(p)$, where $p$ is set as 0.2, similar to SpanBERT \citep{joshi-etal-2020-spanbert}.
All tokens in the selected spans are replaced with either a special $[MASK]$ token, a random token or the original token with probability $80\%$, $10\%$ and $10\%$, respectively.
The goal is to predict these masked tokens $w_m$ based on their surrounding context $w_{\backslash m}$, by minimizing the negative log-likelihood:
\begin{equation}
\begin{aligned}
    \mathcal{L}_{Bidirectional} = - \mathbb{E}_{W \in D} log P_{\theta} (w_m|w_{\backslash m})
\end{aligned}
\label{eq2}
\end{equation}

\paragraph{Seq2Seq generation.} 
For the Seq2Seq generation task, we iteratively sample fragments from the token sequence until the $25\%$ budget has been spent, inspired by \citet{xiao2020ernie}.
For each iterate, we first sample a fragment length from a uniform distribution $l \sim U(4,32)$, and then sample a fragment with the specified length.
Every selected fragment $\{w_i,...,w_j\}$ is further appended with two special tokens $[CLS]$ and $[SEP]$ (i.e., $\{[CLS],w_i,...,w_j,[SEP]\}$), which denotes the beginning and end of the fragment.
All selected fragments are removed from the text and concatenated as the target sequence $T$ while the remaining parts are concatenated as the source sequence $S$.
The model is trained to generate the target sequence auto-regressively condition on the source sequence:
\begin{equation}
\begin{aligned}
    \mathcal{L}_{Seq2Seq} = - \mathbb{E}_{(S,T) \in D} log P_{\theta} (T|S)
\end{aligned}
\label{eq3}
\end{equation}
where $P_{\theta} (T|S)= \prod_{j=1}^{|T|} P_{\theta}(T_j|T_{<j},S)$. During pre-training, we alternate between the bidirectional prediction objective and the Seq2Seq generation objective uniformly.
For image-text pairs, the two objectives are applied to the captions similarly to learn cross-modal understanding and generation.

\section{Experimental Settings}
\label{sec:set}

In this section, we introduce the pre-training and finetuning experimental settings.

\subsection{Pre-training Dataset}
\label{ssec:data}

Our pre-training datasets consist of three types: text corpus, image collections and image-text pairs. 
The text corpus includes two large-scale corpora: BookWiki and OpenWebText, which are part of the training dataset of RoBERTa. BookWiki is composed of English Wikipedia and BookCorpus \citep{zhu2015aligning}, and OpenWebText is an open recreation of the WebText corpora.
The image collections are images without textual descriptions, including a subset of OpenImages \citep{krasin2017openimages} and COCO unlabel.
The image-text pairs are composed of four existing multi-modal datasets: COCO \citep{lin2014microsoft}, Visual Genome (VG) \citep{krishna2017visual}, Conceptual Captions (CC) \citep{sharma-etal-2018-conceptual} and SBU Captions \citep{ordonez2011im2text}, which have also been widely used in previous multi-modal pre-training models.
The statistics of them are shown in Appendix \ref{sec:pretrain}.

\subsection{Implementation Detail}
\label{ssec:imp}
We evaluate UNIMO on two model sizes: UNIMO-base with 12 layers of Transformer block and UNIMO-large with 24 layers of Transformer block. 
The maximum sequence length of text tokens and image-region features are set as 512 and 100, respectively. 
We pre-train UNIMO-base by initializing from RoBERTa-base, and UNIMO-large by initializing from RoBERTa-large. 
Both UNIMO-base and UNIMO-large are trained for at least 500K steps.
An Adam optimizer with initial learning rate 5e-5 and a learning rate linear decay schedule is utilized.
By virtue of float16 mixed precision training, it takes almost 7 days for training UNIMO-base with 32 Nvidia Telsa V100 32GB GPU and 10 days for UNIMO-large with 64 Nvidia Telsa V100 32GB GPU.

For visual learning, we adopt Faster R-CNN \citep{ren2016faster} pre-trained on the Visual-Genome dataset to select salient image regions and extract region features from images.
The regions with class detection probability exceeds a confidence threshold of 0.2 are selected and 100 boxes are kept.
For CMCL, we utilize back-translation to create 3 positive samples and apply rewriting to obtain 100 hard negative samples for each image-text pair.
The most similar of 100 images and 100 sentences are retrieved from the single-modal image collections and text corpus for each image-text pair, respectively.
More details are described in Appendix \ref{sec:pretrain}.

\begin{table*}[ht!]
\centering
\small
\begin{tabular}{l c c c c c}
\hline \hline
\multirow{2}{*}{Model} & Flickr30k-IR & Flickr30k-TR & SNLI-VE & VQA & CoCo Caption  \\
& R@1 / R@5 / R@10 & R@1 / R@5 / R@10 & Val / Test & test-dev / -std  & BLUE4 / CIDEr \\
\hline
ViLBERT-base & 58.20 / 84.90 / 91.52 & - & - & 70.55 / 70.92 & -  \\
VLP-base & - & - & - & 70.5 / 70.7 & 36.5 / 116.9  \\
UNITER-base & 72.52 / 92.36 / \textbf{96.08} & 85.90 / 97.10 / 98.80 & 78.59 / 78.28 & 72.70 / 72.91 & -   \\
Oscar-base & - & - & - & 73.16 / 73.44 & 36.5 / 123.7  \\
Villa-base & \textbf{74.74} / 92.86 / 95.82 & 86.60 / 97.90 / \textbf{99.20} & 79.47 / 79.03 & 73.59 / 73.67 & -  \\
Ernie-ViL-base & 74.44 / 92.72 / 95.94 & 86.70 / 97.80 / 99.00 & - & 72.62 / 72.85 &  -   \\
\textbf{UNIMO-base} & 74.66 / \textbf{93.40} / \textbf{96.08} & \textbf{89.70} / \textbf{98.40} / 99.10 & \textbf{80.00} / \textbf{79.10} & \textbf{73.79} / \textbf{74.02} & \textbf{38.8} / \textbf{124.4}  \\
\hline
UNITER-large & 75.56 / 94.08 / 96.76 & 87.30 / 98.00 / 99.20 & 79.39 / 79.38 & 73.82 / 74.02 &  -   \\
Oscar-large & - & - & - & 73.61 / 73.82 & 37.4 / \textbf{127.8}  \\
Villa-large & 76.26 / \textbf{94.24} / 96.84 & 87.90 / 97.50 / 98.80 & 80.18 / 80.02 & 74.69 / 74.87 & - \\
ERNIE-ViL-large & 76.70 / 93.58 / 96.44 & 88.10 / 98.00 / 99.20 &  - & 74.75 / 74.93 & -  \\
\textbf{UNIMO-large} & \textbf{78.04} / \textbf{94.24} / \textbf{97.12} & \textbf{89.40} / \textbf{98.90} / \textbf{99.80} & \textbf{81.11} / \textbf{80.63} & \textbf{75.06} / \textbf{75.27} & \textbf{39.6} / 127.7 \\
\hline
 \hline
\end{tabular}
\caption{\label{multi-modal}
Evaluation results on the multi-modal downstream tasks.}
\end{table*}

\begin{table*}[ht!]
 \centering
 \setlength{\tabcolsep}{2.7pt}
 \small
 \begin{tabular}{l c c c c c c c c}
  \hline \hline
  \multirow{2}{*}{Model} & SST-2 & MNLI & CoLA & STS-B & CoQA & SQuAD-QG & CNNDM & Gigaword\\
  & Acc & Acc-(m/mm)& Mat & Per & Acc & B4/ME/R-L & R-1/2/L & R-1/2/L \\
  \hline
  BERT-base & 92.7 & 84.4 / - & - & -  & - & - & - & - \\
  RoBERTa-base & 94.8 & - & 63.6 & - & 77.4 & 22.15/24.58/51.12 & 42.31/20.04/39.49 & 38.65/19.66/36.04\\
  \textbf{UNIMO-base} & \textbf{95.1}&\textbf{86.8}/\textbf{86.7} & \textbf{65.4} & \textbf{91.0} & \textbf{80.2} & \textbf{22.78}/\textbf{25.24}/\textbf{51.34} & \textbf{42.42}/\textbf{20.12}/\textbf{39.61} & \textbf{38.80}/\textbf{19.99}/\textbf{36.27} \\
  \ \ w/o single-modal & 82.0 & 59.9/64.9 & 15.0 & 88.8 & 67.1 & 17.09/21.04/46.47 & 41.06/19.01/38.23 & 38.06/18.91/35.41 \\
  \hline
  BERT-large & 93.2 & 86.6/- & 60.6 & 90.0 & - & - & - & - \\
  RoBERTa-large & 96.4 & \textbf{90.2}/\textbf{90.2} & 68.0 & 92.4 & \textbf{85.1} & 23.39/25.73/52.11  & 43.10/20.29/40.24 & 39.32/20.01/36.58 \\
  XLNet-large & 95.6 & 89.8/- & 63.6 & 91.8 & - & - & - & -\\
  UniLM-large & 94.5 & 87.0/85.9 &  61.1 & 87.7 & 82.5 & 22.12/25.06/51.07 & 43.33/20.21/40.51 & 38.45/19.45/35.75 \\
  \textbf{UNIMO-large} & \textbf{96.8} & 89.8/89.5 & \textbf{68.5} & \textbf{92.6} & 84.9 & \textbf{24.59}/\textbf{26.39}/\textbf{52.47}  & \textbf{43.51}/\textbf{20.65}/\textbf{40.63} & \textbf{39.71}/\textbf{20.37}/\textbf{36.88}\\
  \hline
   \hline
 \end{tabular}
 \caption{\label{single-modal}
 Comparison on the single-modal downstream tasks. R-1, R-2 and R-L denote ROUGE-1, ROUGE-2 and ROUGE-L, respectively. Mat, Per, B4 and ME denote Matthews correlation coefficient, Pearson correlation coefficient, BLUE4 and METEOR \citep{lavie2007meteor}, respectively. ``w/o single-modal'' denotes removing the single-modal learning process on the single-modal data from UNIMO, which is similar to UNITER-base \citep{chen2020uniter}. The results on SST-2, MNLI, CoLA, STS-B and CoQA are evaluated on the dev set. The results of RoBERTa on the generation tasks CoQA, SQuAD-QG, CNNDM and Gigaword are evaluated by utilizing the UNIMO architecture initialized with pre-trained parameters of RoBERTa.
 }
\end{table*}

\subsection{Finetuning Tasks}
\label{ssec:fin}

We fine-tune our model on two categories of downstream tasks: (1) single-modal language understanding and generation tasks; (2) multi-modal vision-language understanding and generation tasks. 
The single-modal generation tasks include: 
generative conversational question answering on the CoQA dataset \citep{reddy-etal-2019-coqa},
question generation on the SQuAD 1.1 dataset \citep{rajpurkar-etal-2016-squad},
abstractive summarization on the CNN/DailyMail (CNNDM) dataset \citep{hermann2015teaching}, 
and sentence compression on the Gigaword dataset \citep{rush-etal-2015-neural}.
The single-modal understanding tasks include: sentiment classification on the SST-2 dataset \citep{socher-etal-2013-recursive}, natural language inference on the MNLI dataset \citep{williams2017broad}, linguistic acceptability analysis on the CoLA dataset \citep{warstadt2019cola} and semantic similarity analysis on the STS-B dataset \citep{cer-etal-2017-semeval}.
The multi-modal tasks include: visual question answering (VQA) on the VQA v2.0 dataset \citep{goyal2017making}, image caption on the Microsoft COCO Captions dataset \citep{chen2015microsoft}, visual entailment on the SNLI-VE dataset \citep{xie2019visual} and image-text retrieval on Flickr30k datasets \citep{young-etal-2014-image}. The detail statistics of the datasets and hyper-parameter settings for the above tasks are described in Appendix \ref{sec:finetune}.

\section{Results and Analysis}
\label{sec:res}

In this section, we report the evaluation results on both the multi-modal and single-modal tasks to show the adaptability and generalizability of UNIMO to different scenarios.
We further make several ablation studies to validate that textual knowledge and visual knowledge can enhance each other in the unified semantic space.
The visualization and case analysis of the model results are  appended in Appendix \ref{sec:visualization}.

\subsection{Multi-Modal tasks}
\label{ssec:mlu}

The evaluation results on the multi-modal tasks are shown in Table \ref{multi-modal}.
We compare with most of the existed multi-modal pre-training models, including ViLBERT \citep{lu2019vilbert}, VLP \citep{zhou2020unified}, UNITER \citep{chen2020uniter}, Oscar \citep{li2020oscar}, Villa \citep{gan2020large} and ERNIE-ViL \citep{yu2020ernie}.
The results show that UNIMO achieves the best results against almost all benchmarks under both the base and large size of models.
Particularly, UNIMO-large outperforms previous best performing model ERNIE-ViL-large by 1.34 R@1 on image retrieval and 1.3 R@1 on text retrieval, which are great improvements for the image-text retrieval tasks.
On the image caption task, UNIMO outperforms the best performing model Oscar by more than 2 BLUE4 score.
UNIMO achieves better performance on both the multi-modal understanding and generation tasks, while previous methods usually focus on either the understanding or generation tasks.
The above results demonstrate the effectiveness of the unified-modal learning architecture that takes advantage of the large scale of single-modal images and texts for cross-modal learning.

\begin{table*}[ht!]
\centering
\small
\begin{tabular}{l c c c c c}
\hline \hline
\multirow{2}{*}{Model} & Flickr30k-IR & Flickr30k-TR & SNLI-VE & VQA & CoCo Caption  \\
& R@1 / R@5 / R@10 & R@1 / R@5 / R@10 & Val & test-dev  & BLUE4 / CIDEr \\
\hline
\hline
\textbf{UNIMO-base} & \textbf{74.66} / \textbf{93.40} / \textbf{96.08} & \textbf{89.70} / \textbf{98.40} / 99.10 & \textbf{80.00} & \textbf{73.79} & \textbf{38.8} / \textbf{124.4}  \\
\ \ w/o texts & 72.04 / 91.62 / 95.30 & 85.80 / 97.90 / 99.10 & 79.52 & 73.77 & 38.3 / 123.2 \\
\hline
\end{tabular}
\caption{\label{text-vision}
Analyzing the effectiveness of textual knowledge to multi-modal tasks.
}
\end{table*}

\begin{table*}[ht!]
 \centering
 \setlength{\tabcolsep}{2.7pt}
 \small
 \begin{tabular}{l c c c c c c c c}
  \hline \hline
  \multirow{2}{*}{Model} & SST-2 & MNLI & CoLA & STS-B & CoQA & SQuAD-QG & CNNDM & Gigaword\\
  & Acc & Acc-(m/mm)& Mat & Per & Acc & B4/ME/R-L & R-1/2/L & R-1/2/L \\
  \hline
  \hline
  \textbf{UNIMO-base} & \textbf{95.1}& 86.8/86.7 & \textbf{65.4} & \textbf{91.0} & \textbf{80.2} & \textbf{22.78}/\textbf{25.24}/\textbf{51.34} & \textbf{42.42}/\textbf{20.12}/\textbf{39.61} & \textbf{38.80}/\textbf{19.99}/\textbf{36.27} \\
  w/o pairs\&images & 94.7 & \textbf{87.4}/\textbf{86.8} & 62.8 & 90.6 & 78.1 & 21.26/24.02/50.04 & 42.26/20.09/39.41 & 38.22/19.43/35.71   \\
  \hline
 \end{tabular}
 \caption{\label{vision-text}
Analyzing the effectiveness of visual knowledge to language tasks.
}
\end{table*}

\subsection{Single-Modal tasks}
\label{ssec:sin}

Previous multi-modal pre-training models usually cannot effectively adapt to single-modal scenarios.
To further validate that, we remove the single-modal learning processes on the text corpus and image collections (i.e., ``w/o single-modal'') from UNIMO and replace the CMCL with an image-text matching objective. 
Then, the model ``w/o single-modal'' is just a multi-modal pre-training method similar to UNITER \citep{chen2020uniter}.
As shown in Table \ref{single-modal}, the performance of the model on all the language understanding and generation tasks drop dramatically compared to UNIMO, which demonstrates that multi-modal pre-training only on image-text pairs cannot effectively adapt to the single-modal tasks.

To show the effectiveness of UNIMO on the language understanding and generation tasks, we further compare with existed pre-trained language models (PLMs), including BERT \citep{devlin-etal-2019-bert}, RoBERTa \citep{liu2019roberta}, XLNet \citep{yang2019xlnet} and UniLM \citep{dong2019unified}. 
The comparison results in Table \ref{single-modal} demonstrate that UNIMO achieves better or comparable performance than existed PLMs on both the language understanding and generation tasks.
Specifically, UniLM \citep{dong2019unified} is designed for both natural language understanding and generation.
UNIMO outperforms UniLM on most of the tasks with a large margin, which demonstrates the effectiveness of UNIMO on the single-modal scenarios.

In all, UNIMO not only achieves the best performance on the multi-modal tasks, but also performs very well on the single-modal tasks, which demonstrate the superiority of our unified-modal learning architecture.

\subsection{Mutual Enhancement of Text and Vision}
We further make several ablation studies to show that the unified-modal architecture can help textual knowledge and visual knowledge mutually enhance each other in the unified semantic space.

\paragraph{Text Enhance Vision}
To explore whether the textual knowledge in the text corpus facilitates the cross-modal learning, we remove the language learning process on the text corpus from UNIMO (i.e., ``w/o texts''), and compare their performance on the multi-modal tasks.
Table \ref{text-vision} summarizes the comparison results, which show that the performance of the model ``w/o texts'' declines consistently on both the multi-modal understanding and generation tasks.
The results demonstrate that the textual knowledge in the text corpus benefit the vision-language tasks by enhancing the cross-modal learning with more textual information.

\paragraph{Vision Enhance Text}
To further validate that the visual knowledge in the image collections and image-text pairs facilitates the language learning, we remove the images and image-text pairs from the pre-training dataset (i.e., ``w/o pairs\&images'') and compare their performance on the single-modal language tasks.
After removing the images and image-text pairs, our model is trained by only the language learning objectives, which are similar to previous pre-trained language models BERT and UniLM.
Table \ref{vision-text} summarizes the comparison results, which demonstrate that after removing the visual data, the performance of the model ``w/o pairs\&images'' drops obviously on most of the language understanding tasks and all the language generation tasks.
The results reveal that visual knowledge can enhance the language tasks by enabling the model to learn more robust and generalizable representations in a unified semantic space.

\section{Related Work}
\label{sec:rel}
Existing researches on pre-training can be mainly classified into two categories: single-modal pre-training and multi-modal pre-training.
The single-modal pre-training methods only focus on single-modal tasks, while the multi-modal pre-training methods only focus on multi-modal tasks.

\paragraph{Single-Modal Pre-training}
The single-modal pre-training methods mainly consist of visual pre-training and language pre-training.
Most visual pre-training methods are based on the multi-layer CNN architecture such as VGG \citep{simonyan2014very} and ResNet \citep{he2016deep}, and trained on the ImageNet dataset.
Recently, contrastive self-supervised learning like SimCLR \citep{chen2020simple} and MoCo \citep{he2020momentum} also greatly improve the performance of visual representation learning.
These pre-trained models only focus on visual tasks (e.g. image classification etc.), however, they cannot be used in textual or multi-modal (i.e., with both text and image) tasks.
The language pre-training methods based on the Transformer architecture are also very popular in NLP models, such as GPT \citep{radford2018improving}, BERT \citep{devlin-etal-2019-bert}, XLNet \citep{yang2019xlnet} and BART \citep{lewis-etal-2020-bart}.
However, they mainly focus on textual tasks.
They cannot effectively deal with the multi-modal tasks, such as image-text retrieval, image captioning, multimodal machine translation \citep{lin2020dynamic,su2021multi} and visual dialog \citep{murahari2020large}.

\paragraph{Multi-Modal Pre-training}
Recently, multi-modal pre-training methods have been more and more popular for solving the multi-modal tasks.
All of them are trained on a corpus of image-text pairs, such as ViLBERT \citep{lu2019vilbert}, VisualBERT \citep{li2019visualbert}, VL-BERT \citep{su2019vl}, Unicoder-VL \citep{li2019unicoder} and UNITER \citep{chen2020uniter}.
Based on the multi-layer Transformer network, they all employ the BERT-like objectives to learn multi-modal representations from a concatenated-sequence of vision features and language embeddings.
Their architectures can be mainly classified into two categories: single-stream and two-stream.
The two-stream methods, such as ViLBERT, utilize two single-modal Transformer to process visual features and language embeddings respectively, and then learn their interactions based on a cross-modal Transformer.
The single-stream methods directly utilize a single Transformer network to model both the visual features and the language embeddings. 
VisualBERT, VL-BERT, Unicoder-VL and UNITER all utilize the single-stream architecture, which show that fusing cross-modal information early and freely by a single-stream network can achieve better performance.

Recently, several contrastive learning-based multi-modal pre-training methods have also been proposed. 
OpenAI CLIP \citep{radford2021learning} leverages large-scale image-text pairs to learn transferrable visual representations by image-text matching, which enables zero-shot transfer of the model to various visual classification tasks. 
WenLan \citep{huo2021wenlan} further proposes a similar two-tower Chinese multi-modal pre-training model and adapts MoCo \citep{he2020momentum} to improve the contrastive cross-modal learning process.
Instead of extracting salient image regions by pre-trained object detection models like Faster-RCNN \citep{ren2016faster}, the end-to-end vision-language pre-training architecture SOHO \citep{huang2021seeing} proposes to jointly learn Convolutional Neural Network (CNN) and Transformer for  cross-modal alignments from millions of image-text pairs. 

All existed multi-modal pre-training methods only focus on multi-modal tasks with both vision and language inputs.
However, they cannot be effectively adapted to single-modal tasks.
Moreover, they can only utilize the limited corpus of image-text pairs.
By contrast, our unified-modal pre-training method UNIMO can employ large volumes of text corpus and image collections to enhance each other, and can be effectively adapted to both textual and multi-modal scenarios.
UNIMO also achieves the best performance on multi-modal tasks including image-text retrieval, visual entailment, VQA and image caption.

\section{Conclusion}


In this work, we propose UNIMO, a unified-modal pre-training architecture to leverage the large scale of non-paired text corpus and image collections for cross-modal learning.
We verify that UNIMO provides an effective way for textual knowledge and visual knowledge to mutually enhance each other in a unified semantic space, and UNIMO successfully adapts to both single-modal and multi-modal understanding and generation tasks. 
In this way, UNIMO outperforms previous methods on both the multi-modal and single-modal downstream tasks.
In the future work, we will focus on end-to-end visual and language unified learning, and much larger scale of model size and data volumes.

\section*{Acknowledgments}
This work was supported by the National Key Research and Development Project of China (No. 2018AAA0101900)

\bibliography{acl2021}

\begin{thebibliography}{53}
\expandafter\ifx\csname natexlab\endcsname\relax\def\natexlab#1{#1}\fi

\bibitem[{Cer et~al.(2017)Cer, Diab, Agirre, Lopez-Gazpio, and
  Specia}]{cer-etal-2017-semeval}
Daniel Cer, Mona Diab, Eneko Agirre, I{\~n}igo Lopez-Gazpio, and Lucia Specia.
  2017.
\newblock \href {https://doi.org/10.18653/v1/S17-2001} {{S}em{E}val-2017 task
  1: Semantic textual similarity multilingual and crosslingual focused
  evaluation}.
\newblock In \emph{Proceedings of the 11th International Workshop on Semantic
  Evaluation ({S}em{E}val-2017)}, pages 1--14, Vancouver, Canada. Association
  for Computational Linguistics.

\bibitem[{Chen et~al.(2020{\natexlab{a}})Chen, Kornblith, Norouzi, and
  Hinton}]{chen2020simple}
Ting Chen, Simon Kornblith, Mohammad Norouzi, and Geoffrey Hinton.
  2020{\natexlab{a}}.
\newblock A simple framework for contrastive learning of visual
  representations.
\newblock In \emph{International conference on machine learning}, pages
  1597--1607. PMLR.

\bibitem[{Chen et~al.(2015)Chen, Fang, Lin, Vedantam, Gupta, Doll{\'a}r, and
  Zitnick}]{chen2015microsoft}
Xinlei Chen, Hao Fang, Tsung-Yi Lin, Ramakrishna Vedantam, Saurabh Gupta, Piotr
  Doll{\'a}r, and C~Lawrence Zitnick. 2015.
\newblock Microsoft coco captions: Data collection and evaluation server.
\newblock \emph{arXiv preprint arXiv:1504.00325}.

\bibitem[{Chen et~al.(2020{\natexlab{b}})Chen, Li, Yu, El~Kholy, Ahmed, Gan,
  Cheng, and Liu}]{chen2020uniter}
Yen-Chun Chen, Linjie Li, Licheng Yu, Ahmed El~Kholy, Faisal Ahmed, Zhe Gan,
  Yu~Cheng, and Jingjing Liu. 2020{\natexlab{b}}.
\newblock Uniter: Universal image-text representation learning.
\newblock In \emph{European Conference on Computer Vision}, pages 104--120.
  Springer.

\bibitem[{Devlin et~al.(2019)Devlin, Chang, Lee, and
  Toutanova}]{devlin-etal-2019-bert}
Jacob Devlin, Ming-Wei Chang, Kenton Lee, and Kristina Toutanova. 2019.
\newblock \href {https://doi.org/10.18653/v1/N19-1423} {{BERT}: Pre-training of
  deep bidirectional transformers for language understanding}.
\newblock In \emph{Proceedings of the 2019 Conference of the North {A}merican
  Chapter of the Association for Computational Linguistics: Human Language
  Technologies, Volume 1 (Long and Short Papers)}, pages 4171--4186,
  Minneapolis, Minnesota. Association for Computational Linguistics.

\bibitem[{Dong et~al.(2019)Dong, Yang, Wang, Wei, Liu, Wang, Gao, Zhou, and
  Hon}]{dong2019unified}
Li~Dong, Nan Yang, Wenhui Wang, Furu Wei, Xiaodong Liu, Yu~Wang, Jianfeng Gao,
  Ming Zhou, and Hsiao-Wuen Hon. 2019.
\newblock Unified language model pre-training for natural language
  understanding and generation.
\newblock In \emph{Advances in Neural Information Processing Systems}, pages
  13063--13075.

\bibitem[{Edunov et~al.(2018)Edunov, Ott, Auli, and
  Grangier}]{edunov-etal-2018-understanding}
Sergey Edunov, Myle Ott, Michael Auli, and David Grangier. 2018.
\newblock \href {https://doi.org/10.18653/v1/D18-1045} {Understanding
  back-translation at scale}.
\newblock In \emph{Proceedings of the 2018 Conference on Empirical Methods in
  Natural Language Processing}, pages 489--500, Brussels, Belgium. Association
  for Computational Linguistics.

\bibitem[{Gan et~al.(2020)Gan, Chen, Li, Zhu, Cheng, and Liu}]{gan2020large}
Zhe Gan, Yen-Chun Chen, Linjie Li, Chen Zhu, Yu~Cheng, and Jingjing Liu. 2020.
\newblock Large-scale adversarial training for vision-and-language
  representation learning.
\newblock \emph{arXiv preprint arXiv:2006.06195}.

\bibitem[{Goyal et~al.(2017)Goyal, Khot, Summers-Stay, Batra, and
  Parikh}]{goyal2017making}
Yash Goyal, Tejas Khot, Douglas Summers-Stay, Dhruv Batra, and Devi Parikh.
  2017.
\newblock Making the v in vqa matter: Elevating the role of image understanding
  in visual question answering.
\newblock In \emph{Proceedings of the IEEE Conference on Computer Vision and
  Pattern Recognition}, pages 6904--6913.

\bibitem[{He et~al.(2020)He, Fan, Wu, Xie, and Girshick}]{he2020momentum}
Kaiming He, Haoqi Fan, Yuxin Wu, Saining Xie, and Ross Girshick. 2020.
\newblock Momentum contrast for unsupervised visual representation learning.
\newblock In \emph{Proceedings of the IEEE/CVF Conference on Computer Vision
  and Pattern Recognition}, pages 9729--9738.

\bibitem[{He et~al.(2016)He, Zhang, Ren, and Sun}]{he2016deep}
Kaiming He, Xiangyu Zhang, Shaoqing Ren, and Jian Sun. 2016.
\newblock Deep residual learning for image recognition.
\newblock In \emph{Proceedings of the IEEE conference on computer vision and
  pattern recognition}, pages 770--778.

\bibitem[{Hermann et~al.(2015)Hermann, Kocisky, Grefenstette, Espeholt, Kay,
  Suleyman, and Blunsom}]{hermann2015teaching}
Karl~Moritz Hermann, Tomas Kocisky, Edward Grefenstette, Lasse Espeholt, Will
  Kay, Mustafa Suleyman, and Phil Blunsom. 2015.
\newblock Teaching machines to read and comprehend.
\newblock \emph{Advances in neural information processing systems},
  28:1693--1701.

\bibitem[{Huang et~al.(2021)Huang, Zeng, Huang, Liu, Fu, and
  Fu}]{huang2021seeing}
Zhicheng Huang, Zhaoyang Zeng, Yupan Huang, Bei Liu, Dongmei Fu, and Jianlong
  Fu. 2021.
\newblock Seeing out of the box: End-to-end pre-training for vision-language
  representation learning.
\newblock \emph{arXiv preprint arXiv:2104.03135}.

\bibitem[{Huo et~al.(2021)Huo, Zhang, Liu, Lu, Gao, Yang, Wen, Zhang, Xu, Zheng
  et~al.}]{huo2021wenlan}
Yuqi Huo, Manli Zhang, Guangzhen Liu, Haoyu Lu, Yizhao Gao, Guoxing Yang,
  Jingyuan Wen, Heng Zhang, Baogui Xu, Weihao Zheng, et~al. 2021.
\newblock Wenlan: Bridging vision and language by large-scale multi-modal
  pre-training.
\newblock \emph{arXiv preprint arXiv:2103.06561}.

\bibitem[{Joshi et~al.(2020)Joshi, Chen, Liu, Weld, Zettlemoyer, and
  Levy}]{joshi-etal-2020-spanbert}
Mandar Joshi, Danqi Chen, Yinhan Liu, Daniel~S. Weld, Luke Zettlemoyer, and
  Omer Levy. 2020.
\newblock \href {https://doi.org/10.1162/tacl_a_00300} {{S}pan{BERT}: Improving
  pre-training by representing and predicting spans}.
\newblock \emph{Transactions of the Association for Computational Linguistics},
  8:64--77.

\bibitem[{Krasin et~al.(2017)Krasin, Duerig, Alldrin, Ferrari, Abu-El-Haija,
  Kuznetsova, Rom, Uijlings, Popov, Veit et~al.}]{krasin2017openimages}
Ivan Krasin, Tom Duerig, Neil Alldrin, Vittorio Ferrari, Sami Abu-El-Haija,
  Alina Kuznetsova, Hassan Rom, Jasper Uijlings, Stefan Popov, Andreas Veit,
  et~al. 2017.
\newblock Openimages: A public dataset for large-scale multi-label and
  multi-class image classification.
\newblock \emph{Dataset available from https://github. com/openimages},
  2(3):2--3.

\bibitem[{Krishna et~al.(2017)Krishna, Zhu, Groth, Johnson, Hata, Kravitz,
  Chen, Kalantidis, Li, Shamma et~al.}]{krishna2017visual}
Ranjay Krishna, Yuke Zhu, Oliver Groth, Justin Johnson, Kenji Hata, Joshua
  Kravitz, Stephanie Chen, Yannis Kalantidis, Li-Jia Li, David~A Shamma, et~al.
  2017.
\newblock Visual genome: Connecting language and vision using crowdsourced
  dense image annotations.
\newblock \emph{International journal of computer vision}, 123(1):32--73.

\bibitem[{Krizhevsky et~al.(2017)Krizhevsky, Sutskever, and
  Hinton}]{krizhevsky2017imagenet}
Alex Krizhevsky, Ilya Sutskever, and Geoffrey~E Hinton. 2017.
\newblock Imagenet classification with deep convolutional neural networks.
\newblock \emph{Communications of the ACM}, 60(6):84--90.

\bibitem[{Lavie and Agarwal(2007)}]{lavie2007meteor}
Alon Lavie and Abhaya Agarwal. 2007.
\newblock Meteor: An automatic metric for mt evaluation with high levels of
  correlation with human judgments.
\newblock In \emph{Proceedings of the second workshop on statistical machine
  translation}, pages 228--231.

\bibitem[{Lewis et~al.(2020)Lewis, Liu, Goyal, Ghazvininejad, Mohamed, Levy,
  Stoyanov, and Zettlemoyer}]{lewis-etal-2020-bart}
Mike Lewis, Yinhan Liu, Naman Goyal, Marjan Ghazvininejad, Abdelrahman Mohamed,
  Omer Levy, Veselin Stoyanov, and Luke Zettlemoyer. 2020.
\newblock \href {https://doi.org/10.18653/v1/2020.acl-main.703} {{BART}:
  Denoising sequence-to-sequence pre-training for natural language generation,
  translation, and comprehension}.
\newblock In \emph{Proceedings of the 58th Annual Meeting of the Association
  for Computational Linguistics}, pages 7871--7880, Online. Association for
  Computational Linguistics.

\bibitem[{Li et~al.(2019{\natexlab{a}})Li, Duan, Fang, Jiang, and
  Zhou}]{li2019unicoder}
Gen Li, Nan Duan, Yuejian Fang, Daxin Jiang, and Ming Zhou. 2019{\natexlab{a}}.
\newblock Unicoder-vl: A universal encoder for vision and language by
  cross-modal pre-training.
\newblock \emph{arXiv preprint arXiv:1908.06066}.

\bibitem[{Li et~al.(2019{\natexlab{b}})Li, Yatskar, Yin, Hsieh, and
  Chang}]{li2019visualbert}
Liunian~Harold Li, Mark Yatskar, Da~Yin, Cho-Jui Hsieh, and Kai-Wei Chang.
  2019{\natexlab{b}}.
\newblock Visualbert: A simple and performant baseline for vision and language.
\newblock \emph{arXiv preprint arXiv:1908.03557}.

\bibitem[{Li et~al.(2020)Li, Yin, Li, Zhang, Hu, Zhang, Wang, Hu, Dong, Wei
  et~al.}]{li2020oscar}
Xiujun Li, Xi~Yin, Chunyuan Li, Pengchuan Zhang, Xiaowei Hu, Lei Zhang, Lijuan
  Wang, Houdong Hu, Li~Dong, Furu Wei, et~al. 2020.
\newblock Oscar: Object-semantics aligned pre-training for vision-language
  tasks.
\newblock In \emph{European Conference on Computer Vision}, pages 121--137.
  Springer.

\bibitem[{Lin et~al.(2020{\natexlab{a}})Lin, Meng, Su, Yin, Yang, Ge, Zhou, and
  Luo}]{lin2020dynamic}
Huan Lin, Fandong Meng, Jinsong Su, Yongjing Yin, Zhengyuan Yang, Yubin Ge, Jie
  Zhou, and Jiebo Luo. 2020{\natexlab{a}}.
\newblock Dynamic context-guided capsule network for multimodal machine
  translation.
\newblock In \emph{Proceedings of the 28th ACM International Conference on
  Multimedia}, pages 1320--1329.

\bibitem[{Lin et~al.(2020{\natexlab{b}})Lin, Yang, Zhang, Liu, Zhou, and
  Yang}]{lin2020interbert}
Junyang Lin, An~Yang, Yichang Zhang, Jie Liu, Jingren Zhou, and Hongxia Yang.
  2020{\natexlab{b}}.
\newblock Interbert: Vision-and-language interaction for multi-modal
  pretraining.
\newblock \emph{arXiv preprint arXiv:2003.13198}.

\bibitem[{Lin et~al.(2014)Lin, Maire, Belongie, Hays, Perona, Ramanan,
  Doll{\'a}r, and Zitnick}]{lin2014microsoft}
Tsung-Yi Lin, Michael Maire, Serge Belongie, James Hays, Pietro Perona, Deva
  Ramanan, Piotr Doll{\'a}r, and C~Lawrence Zitnick. 2014.
\newblock Microsoft coco: Common objects in context.
\newblock In \emph{European conference on computer vision}, pages 740--755.
  Springer.

\bibitem[{Liu et~al.(2019)Liu, Ott, Goyal, Du, Joshi, Chen, Levy, Lewis,
  Zettlemoyer, and Stoyanov}]{liu2019roberta}
Yinhan Liu, Myle Ott, Naman Goyal, Jingfei Du, Mandar Joshi, Danqi Chen, Omer
  Levy, Mike Lewis, Luke Zettlemoyer, and Veselin Stoyanov. 2019.
\newblock Roberta: A robustly optimized bert pretraining approach.
\newblock \emph{arXiv preprint arXiv:1907.11692}.

\bibitem[{Lu et~al.(2019)Lu, Batra, Parikh, and Lee}]{lu2019vilbert}
Jiasen Lu, Dhruv Batra, Devi Parikh, and Stefan Lee. 2019.
\newblock Vilbert: Pretraining task-agnostic visiolinguistic representations
  for vision-and-language tasks.
\newblock In \emph{Advances in Neural Information Processing Systems}, pages
  13--23.

\bibitem[{Murahari et~al.(2020)Murahari, Batra, Parikh, and
  Das}]{murahari2020large}
Vishvak Murahari, Dhruv Batra, Devi Parikh, and Abhishek Das. 2020.
\newblock Large-scale pretraining for visual dialog: A simple state-of-the-art
  baseline.
\newblock In \emph{European Conference on Computer Vision}, pages 336--352.
  Springer.

\bibitem[{Ordonez et~al.(2011)Ordonez, Kulkarni, and Berg}]{ordonez2011im2text}
Vicente Ordonez, Girish Kulkarni, and Tamara Berg. 2011.
\newblock Im2text: Describing images using 1 million captioned photographs.
\newblock \emph{Advances in neural information processing systems},
  24:1143--1151.

\bibitem[{Radford et~al.(2021)Radford, Kim, Hallacy, Ramesh, Goh, Agarwal,
  Sastry, Askell, Mishkin, Clark et~al.}]{radford2021learning}
Alec Radford, Jong~Wook Kim, Chris Hallacy, Aditya Ramesh, Gabriel Goh,
  Sandhini Agarwal, Girish Sastry, Amanda Askell, Pamela Mishkin, Jack Clark,
  et~al. 2021.
\newblock Learning transferable visual models from natural language
  supervision.
\newblock \emph{arXiv preprint arXiv:2103.00020}.

\bibitem[{Radford et~al.(2018)Radford, Narasimhan, Salimans, and
  Sutskever}]{radford2018improving}
Alec Radford, Karthik Narasimhan, Tim Salimans, and Ilya Sutskever. 2018.
\newblock Improving language understanding by generative pre-training.

\bibitem[{Rajpurkar et~al.(2016)Rajpurkar, Zhang, Lopyrev, and
  Liang}]{rajpurkar-etal-2016-squad}
Pranav Rajpurkar, Jian Zhang, Konstantin Lopyrev, and Percy Liang. 2016.
\newblock \href {https://doi.org/10.18653/v1/D16-1264} {{SQ}u{AD}: 100,000+
  questions for machine comprehension of text}.
\newblock In \emph{Proceedings of the 2016 Conference on Empirical Methods in
  Natural Language Processing}, pages 2383--2392, Austin, Texas. Association
  for Computational Linguistics.

\bibitem[{Reddy et~al.(2019)Reddy, Chen, and Manning}]{reddy-etal-2019-coqa}
Siva Reddy, Danqi Chen, and Christopher~D. Manning. 2019.
\newblock \href {https://doi.org/10.1162/tacl_a_00266} {{C}o{QA}: A
  conversational question answering challenge}.
\newblock \emph{Transactions of the Association for Computational Linguistics},
  7:249--266.

\bibitem[{Ren et~al.(2016)Ren, He, Girshick, and Sun}]{ren2016faster}
Shaoqing Ren, Kaiming He, Ross Girshick, and Jian Sun. 2016.
\newblock Faster r-cnn: Towards real-time object detection with region proposal
  networks.
\newblock \emph{IEEE transactions on pattern analysis and machine
  intelligence}, 39(6):1137--1149.

\bibitem[{Rush et~al.(2015)Rush, Chopra, and Weston}]{rush-etal-2015-neural}
Alexander~M. Rush, Sumit Chopra, and Jason Weston. 2015.
\newblock \href {https://doi.org/10.18653/v1/D15-1044} {A neural attention
  model for abstractive sentence summarization}.
\newblock In \emph{Proceedings of the 2015 Conference on Empirical Methods in
  Natural Language Processing}, pages 379--389, Lisbon, Portugal. Association
  for Computational Linguistics.

\bibitem[{Sennrich et~al.(2016)Sennrich, Haddow, and
  Birch}]{sennrich-etal-2016-neural}
Rico Sennrich, Barry Haddow, and Alexandra Birch. 2016.
\newblock \href {https://doi.org/10.18653/v1/P16-1162} {Neural machine
  translation of rare words with subword units}.
\newblock In \emph{Proceedings of the 54th Annual Meeting of the Association
  for Computational Linguistics (Volume 1: Long Papers)}, pages 1715--1725,
  Berlin, Germany. Association for Computational Linguistics.

\bibitem[{Sharma et~al.(2018)Sharma, Ding, Goodman, and
  Soricut}]{sharma-etal-2018-conceptual}
Piyush Sharma, Nan Ding, Sebastian Goodman, and Radu Soricut. 2018.
\newblock \href {https://doi.org/10.18653/v1/P18-1238} {Conceptual captions: A
  cleaned, hypernymed, image alt-text dataset for automatic image captioning}.
\newblock In \emph{Proceedings of the 56th Annual Meeting of the Association
  for Computational Linguistics (Volume 1: Long Papers)}, pages 2556--2565,
  Melbourne, Australia. Association for Computational Linguistics.

\bibitem[{Simonyan and Zisserman(2014)}]{simonyan2014very}
Karen Simonyan and Andrew Zisserman. 2014.
\newblock Very deep convolutional networks for large-scale image recognition.
\newblock \emph{arXiv preprint arXiv:1409.1556}.

\bibitem[{Socher et~al.(2013)Socher, Perelygin, Wu, Chuang, Manning, Ng, and
  Potts}]{socher-etal-2013-recursive}
Richard Socher, Alex Perelygin, Jean Wu, Jason Chuang, Christopher~D. Manning,
  Andrew Ng, and Christopher Potts. 2013.
\newblock \href {https://www.aclweb.org/anthology/D13-1170} {Recursive deep
  models for semantic compositionality over a sentiment treebank}.
\newblock In \emph{Proceedings of the 2013 Conference on Empirical Methods in
  Natural Language Processing}, pages 1631--1642, Seattle, Washington, USA.
  Association for Computational Linguistics.

\bibitem[{Su et~al.(2021)Su, Chen, Jiang, Zhou, Lin, Ge, Wu, and
  Lai}]{su2021multi}
Jinsong Su, Jinchang Chen, Hui Jiang, Chulun Zhou, Huan Lin, Yubin Ge,
  Qingqiang Wu, and Yongxuan Lai. 2021.
\newblock Multi-modal neural machine translation with deep semantic
  interactions.
\newblock \emph{Information Sciences}, 554:47--60.

\bibitem[{Su et~al.(2019)Su, Zhu, Cao, Li, Lu, Wei, and Dai}]{su2019vl}
Weijie Su, Xizhou Zhu, Yue Cao, Bin Li, Lewei Lu, Furu Wei, and Jifeng Dai.
  2019.
\newblock Vl-bert: Pre-training of generic visual-linguistic representations.
\newblock \emph{arXiv preprint arXiv:1908.08530}.

\bibitem[{Van~Ackeren et~al.(2018)Van~Ackeren, Barbero, Mattioni, Bottini, and
  Collignon}]{van2018neuronal}
Markus~Johannes Van~Ackeren, Francesca~M Barbero, Stefania Mattioni, Roberto
  Bottini, and Olivier Collignon. 2018.
\newblock Neuronal populations in the occipital cortex of the blind synchronize
  to the temporal dynamics of speech.
\newblock \emph{ELife}, 7:e31640.

\bibitem[{Wang et~al.(2018)Wang, Liu, Zeng, and Yuille}]{wang-etal-2018-scene}
Yu-Siang Wang, Chenxi Liu, Xiaohui Zeng, and Alan Yuille. 2018.
\newblock \href {https://doi.org/10.18653/v1/N18-1037} {Scene graph parsing as
  dependency parsing}.
\newblock In \emph{Proceedings of the 2018 Conference of the North {A}merican
  Chapter of the Association for Computational Linguistics: Human Language
  Technologies, Volume 1 (Long Papers)}, pages 397--407, New Orleans,
  Louisiana. Association for Computational Linguistics.

\bibitem[{Warstadt et~al.(2019)Warstadt, Singh, and Bowman}]{warstadt2019cola}
Alex Warstadt, Amanpreet Singh, and Samuel~R Bowman. 2019.
\newblock Cola: The corpus of linguistic acceptability (with added
  annotations).

\bibitem[{Williams et~al.(2017)Williams, Nangia, and
  Bowman}]{williams2017broad}
Adina Williams, Nikita Nangia, and Samuel~R Bowman. 2017.
\newblock A broad-coverage challenge corpus for sentence understanding through
  inference.
\newblock \emph{arXiv preprint arXiv:1704.05426}.

\bibitem[{Xiao et~al.(2020)Xiao, Zhang, Li, Sun, Tian, Wu, and
  Wang}]{xiao2020ernie}
Dongling Xiao, Han Zhang, Yukun Li, Yu~Sun, Hao Tian, Hua Wu, and Haifeng Wang.
  2020.
\newblock Ernie-gen: An enhanced multi-flow pre-training and fine-tuning
  framework for natural language generation.
\newblock \emph{arXiv preprint arXiv:2001.11314}.

\bibitem[{Xie et~al.(2019)Xie, Lai, Doran, and Kadav}]{xie2019visual}
Ning Xie, Farley Lai, Derek Doran, and Asim Kadav. 2019.
\newblock Visual entailment: A novel task for fine-grained image understanding.
\newblock \emph{arXiv preprint arXiv:1901.06706}.

\bibitem[{Yang et~al.(2019)Yang, Dai, Yang, Carbonell, Salakhutdinov, and
  Le}]{yang2019xlnet}
Zhilin Yang, Zihang Dai, Yiming Yang, Jaime Carbonell, Russ~R Salakhutdinov,
  and Quoc~V Le. 2019.
\newblock Xlnet: Generalized autoregressive pretraining for language
  understanding.
\newblock In \emph{Advances in neural information processing systems}, pages
  5753--5763.

\bibitem[{Young et~al.(2014)Young, Lai, Hodosh, and
  Hockenmaier}]{young-etal-2014-image}
Peter Young, Alice Lai, Micah Hodosh, and Julia Hockenmaier. 2014.
\newblock \href {https://doi.org/10.1162/tacl_a_00166} {From image descriptions
  to visual denotations: New similarity metrics for semantic inference over
  event descriptions}.
\newblock \emph{Transactions of the Association for Computational Linguistics},
  2:67--78.

\bibitem[{Yu et~al.(2020)Yu, Tang, Yin, Sun, Tian, Wu, and Wang}]{yu2020ernie}
Fei Yu, Jiji Tang, Weichong Yin, Yu~Sun, Hao Tian, Hua Wu, and Haifeng Wang.
  2020.
\newblock Ernie-vil: Knowledge enhanced vision-language representations through
  scene graph.
\newblock \emph{arXiv preprint arXiv:2006.16934}.

\bibitem[{Zhou et~al.(2020)Zhou, Palangi, Zhang, Hu, Corso, and
  Gao}]{zhou2020unified}
Luowei Zhou, Hamid Palangi, Lei Zhang, Houdong Hu, Jason Corso, and Jianfeng
  Gao. 2020.
\newblock Unified vision-language pre-training for image captioning and vqa.
\newblock In \emph{Proceedings of the AAAI Conference on Artificial
  Intelligence}, volume~34, pages 13041--13049.

\bibitem[{Zhu et~al.(2015)Zhu, Kiros, Zemel, Salakhutdinov, Urtasun, Torralba,
  and Fidler}]{zhu2015aligning}
Yukun Zhu, Ryan Kiros, Rich Zemel, Ruslan Salakhutdinov, Raquel Urtasun,
  Antonio Torralba, and Sanja Fidler. 2015.
\newblock Aligning books and movies: Towards story-like visual explanations by
  watching movies and reading books.
\newblock In \emph{Proceedings of the IEEE international conference on computer
  vision}, pages 19--27.

\end{thebibliography}
\bibliographystyle{acl_natbib}

\cleardoublepage

\appendix

\begin{table*}[t!]
\centering
\small
\begin{tabular}{l|l|l|l|l|l|l|l}
\hline
Type& \multicolumn{4}{c|}{Image-Text Pairs} & \multirow{2}{*}{Images} & \multicolumn{2}{c}{Text Corpus} \\
\cline{1-5} \cline{7-8}
Dataset & COCO & VG & CC & SBU & & BookWiki & OpenWebText \\
\hline
Train & 533K & 5.06M & 3.0M & 990K & 1.7M & 16G & 38G \\
Val & 25K & 106K & 14K & 10K & & & \\
\hline
\end{tabular}
\caption{\label{pretrain-dataset}
Statistics of the image-text pairs, image collections and text corpus for pre-training.}
\end{table*}

\begin{table}[t!]
\centering
\small
\begin{tabular}{l|l|l}
\hline
Hyper-parameters & UNIMO-Base & UNIMO-Large\\
\hline
Num of Layers & 12 & 24 \\
Hidden Size & 768 & 1024 \\
FFN Hidden Size & 3072 & 4096 \\
Attention Heads & 12 & 16 \\
Head Size & 64 & 64 \\
Dropout & 0.1 & 0.1 \\
Attention Dropout & 0.1 & 0.1 \\
Warmup Steps & 24K & 30K \\
Peak Learning Rate & 5e-5 & 5e-5 \\
Batch Size & 6K & 3K \\
Weight Decay & 0.01 & 0.01 \\
Max Training Steps & 1M & 1M \\
Learning Rate Decay & Linear & Linear \\
Adam $\epsilon$ & 1e-6 & 1e-6 \\
Adam $\beta_1$ & 0.9 & 0.9 \\
Adam $\beta_2$ & 0.999 & 0.999 \\
Gradient Clipping & 1.0 & 1.0 \\
\hline
\end{tabular}
\caption{\label{pre-params}
Hyper-parameters for UNIMO pre-training.}
\end{table}

\section{Pre-training Settings}
\label{sec:pretrain}

\paragraph{Data Processing}
The pre-training datasets consist of text corpus, image collections and image-text pairs.
The detail statistics of them are shown in Table \ref{pretrain-dataset}. 
For unified-modal learning, all data (including images, texts and image-text pairs) are represented in the same format with both visual and textual input as ``[IMG] [box1] ... [box100] [CLS] [tok1] ... [tokN] [SEP]'', which ``[box]'' and ``[tok]'' denote an image region and subword token, respectively.
For \textbf{single-modal images}, a pseudo token sequence ``[CLS] [PAD] ... [SEP]'' is treated as the textual input during pre-training.
During visual learning on images, the pseudo token sequence will be masked out by special self-attention masks to eliminate its effect to the visual learning process.
The language learning process will not be applied on the pseudo token sequence.
So the single-modal images are equivalent to be encoded individually rather than in pair.
Similarly, for \textbf{single-modal texts}, a pseudo image-region sequence ``[IMG] [0] ... [0]'' will be utilized as the visual input, where ``[0]'' denotes a zero-value feature embedding.
During language learning, the pseudo image-region sequence will be masked out.
Based on the above techniques, both images and texts are represented in the same format as image-text pairs.
For \textbf{image-text pairs}, both the visual learning and language learning are applied on the images and captions simultaneously to learn cross-modal representations. 

\paragraph{Training Details} During pre-training, the samples of image collections, text corpus and image-text pairs are randomly mixed together with ratio 1:1:5. The objectives of language learning, visual learning and cross-modal contrastive learning (CMCL) are trained jointly.
The hyper-parameters for both UNIMO-Base and UNIMO-Large are shown in Table \ref{pre-params}.
For CMCL, each positive image-text pair is appended with several hard negative samples by text rewriting, as well as several positive images and texts by image/text retrieval.
All samples for other image-text pairs in the training batch are also treated as the negative samples (including negative images and negative texts), which are more than 6K for UNIMO-base and 3K for UNIMO-Large.
For an image-text pair $(V,W)$, the detail formula of the CMCL loss $\mathcal{L}_{CMCL}(V,W)$ is as follows:
\begin{equation}
\small
\begin{aligned}
      -log \frac{pos_P + pos_I + pos_T} {(neg_P + neg_I + neg_T) + (pos_P + pos_I + pos_T)}
\end{aligned}
\label{eq4}
\end{equation}
\begin{equation}
\small
\begin{aligned}
      \left\{
      \begin{aligned}
      pos_P & = \sum_{(V^{+},W^{+}) \in \mathcal{X}^{+}} exp(d(V^{+},W^{+})/ \tau) \\
      pos_I & = \sum_{V^{r} \in \mathcal{X}^{I}} exp(d(V^{r},W)/ \tau) \\
      pos_T & = \sum_{W^{r} \in \mathcal{X}^{T}} exp(d(V,W^{r})/ \tau) \\
      neg_P & = \sum_{(V^{-}, W^{-}) \in \mathcal{X}^{-}} exp(d(V^{-}, W^{-})/ \tau) \\
      neg_I & = \sum_{V' \in \mathcal{Y}^{I}} exp(d(V', W)/ \tau) \\
      neg_T & = \sum_{W' \in \mathcal{Y}^{T}} exp(d(V, W')/ \tau) \\
      \end{aligned}
      \right.
\end{aligned}
\label{eq41}
\end{equation}
where $pos_P$, $pos_I$ and $pos_T$ denote the scores of positive image-text pairs $\mathcal{X^+}$, related images $\mathcal{X^I}$ and related texts $\mathcal{X^T}$, respectively.
Also, $neg_P$, $neg_I$ and $neg_T$ denote the scores of negative image-text pairs $\mathcal{X^-}$, negative images $\mathcal{Y^I}$ and negative texts $\mathcal{Y^T}$, respectively.
The objective is to maximize the positive score $pos_P + pos_I + pos_T$ while minimizing the negative score $neg_P + neg_I + neg_T$, while help aligns and unifies the visual and textual representation spaces. The pre-training process of UNIMO is described in Algorithm \ref{alg:cmcl-pseudocode} in pseudo-code style.

\paragraph{Data Augmentations} We apply two types of data augmentation techniques in the CMCL: text rewriting and image/text retrieval. The \textbf{text rewriting} techniques are utilized to create positive and negative examples for CMCL. To create more positive image-text pairs, we apply back-translation to all captions in the image-text pairs. Each caption is translated into 3 kinds of languages, including Chinese, French and Spanish, by our translation tool in house, and then translated back to English. For the phrase-level and word-level rewriting, each caption in the image-text pairs is firstly parsed into a scene graph by the Stanford Scene Graph Parser\footnote{https://nlp.stanford.edu/software/scenegraph-parser.shtml}. All objects, attributes and relations are extracted to build an object vocabulary, an attribute vocabulary and a relation vocabulary. For each caption, the objects, attributes or relations are randomly replaced with other similar objects, attributes or relations in the corresponding vocabularies, respectively. The rewritten captions are ranked based on their linguistic fluency, and the top 100 captions are selected to create hard negative image-text pairs by composing with the original image.
Furthermore, the \textbf{image and text retrieval} techniques are utilized to augment each image-text pair with various related images and texts from the single-modal image collections and text corpus.
For image-retrieval, each image is transformed into 100 image regions and the object labels are detected for all regions by Faster R-CNN.
The object labels are utilized to create a TF-IDF feature vector for each image, and the cosine similarity between images are computed.
For each image in the image-text pairs, 100 of the most similar images are retrieved from the image collections, which are treated as positive images in the CMCL.
For text retrieval, we firstly build an inverted index for all image captions and sentences in the text corpus, then filter non-relevant sentences from the text corpus based on the inverted index.
For each caption in the image-text pairs, the TF-IDF similarities between the caption and the relevant sentences retrieved by the inverted index are calculated, and the top-1000 sentences are extracted.
Further, BERT-based embedding similarities are computed between the caption and the 1000 sentences to rank them, and the top-100 sentences are extracted as the positive texts for the CMCL.

\begin{table*}[t!]
\centering
\small
\begin{tabular}{l|l|l|l|l|l}
\hline
\multirow{3}{*}{Task} & \multirow{3}{*}{Image Src.} & \multicolumn{4}{c}{\#Images (\#Text)} \\
\cline{3-6}
&& \multirow{2}{*}{Train} & \multirow{2}{*}{Val} & \multicolumn{2}{c}{Test} \\
\cline{5-6}
&&&& test-std & test-dev \\
\hline
VQA & COCO & 83K(444K) & 41K(214K) & 81K(107K) & 81K(448K) \\
\hline
Image Caption & COCO & 113.2K & 5K & 5K & - \\
\hline
Visual Entailment & Flickr30K & 529.5K & 17.9K & 17.9K & - \\
\hline
Image-Text Retrieval & Flickr30K & 29K(145K) & 1K(5K) & 1K(5K) & - \\
\hline
\end{tabular}
\caption{\label{finetune-dataset}
Statistics of the datasets for the multi-modal downstream tasks.}
\end{table*}

\begin{table*}[t!]
\centering
\small
\begin{tabular}{l|l|l|l|l}
\hline
Hyper-parameters & Image-Text Retrieval & SNLI-VE & VQA & COCO Caption \\
\hline
\hline
Batch Size & 64/32 & 192/64 & 256/256 & 64/32 \\
\hline
Epoch & 40 & 10 & 12 & 10 \\
\hline
\multirow{3}{*}{Learning Rate} & 5e-6 for epoch=[0,24] & \multirow{3}{*}{1e-5} & 1e-4/4e-5 for epoch=[0,5] & 1e-5/5e-6 \\
& 5e-7 for epoch=[24,32] & & 1e-5/4e-6 for epoch=[6,8] & \\
& 5e-8 for epoch=[32,40] & & 1e-6/4e-7 for epoch=[9,12] & \\
\hline
Warmup Ratio & - & 0.06 & - & 0.06 \\
\hline
Weight Decay & 0.01 & 0.0 & 0.01 & 0.01 \\
\hline
\end{tabular}
\caption{\label{multimodal-params}
Hyper-parameters (base/large) for fine-tuning multi-modal tasks .}
\end{table*}

\begin{table*}[t!]
\centering
\small
\begin{tabular}{l|c|c|c|c|c}
\hline
Hyper-parameters & SST-2/MNLI/CoLA/STS-B & CNNDM & Gigaword & SQuAD-QG & CoQA \\
\hline
Learning Rate & \{1e-5, 2e-5, 3e-5\} & 4e-5/2e-5 & 3e-5 & 1.25e-5/5e-6 & 1e-5/8e-6 \\
Batch Size & \{16, 32\} & 32 & 128 & 32 & 32 \\
Epochs & 10 & 20 & 10 & 20 & 20 \\
Warmup Raito & 0.06 & 0.06 & 0.06 & 0.06 & 0.06 \\
Beam Size & - & 6 & 6 & 6 & 3 \\
Length Penalty & - & 0.6/1.2 & 0.6/1.2 & 1.0/1.2 & 0.0 \\
Trigram Blocking & - &True & False & False & False \\
\hline
\end{tabular}
\caption{\label{gen-params}
Hyper-parameters (base/large) for fine-tuning single-modal tasks.}
\end{table*}

\begin{algorithm}[t!]  
	\caption{UNIMO's pre-training process in a Python-like style.}  
	\label{alg:cmcl-pseudocode} 
	\begin{algorithmic}
	\small{
		\State \leftline{\color{gray}{\# \texttt{\scriptsize{The training details of UNIMO}}}}
		\Function {$\mathbf{pretraining\_ process}$}{}
		\For{$step$ in $all\_steps$}
		\State $batch = []$
		\State \leftline{\color{gray}{\# \texttt{\scriptsize{load $x$ image samples}}}}
		\State $imgs = get\_data(ImgCollections,\ x)$
		\State \leftline{\color{gray}{\# \texttt{\scriptsize{load $y$ text samples}}}}
		\State $texts = get\_data(TextCorpus,\ y)$
		\State \leftline{\color{gray}{\# \texttt{\scriptsize{load $z$ image-text pairs}}}}
		\State $img\_text\_pairs = get\_data(Pairs,\ z)$
		\State \leftline{\color{gray}{\# \texttt{\scriptsize{load CMCL data for each image-text pair}}}}
		\For{$pair$ in $img\_text\_pairs$}
		\State$samples = \mathbf{cmcl\_data\_loader}(pair)$
		\State$batch.extend(samples)$
		\EndFor
		\State$batch.extend(texts)$
		\State$batch.extend(imgs)$
		\State $v\_loss,\ l\_loss,\ cmcl\_loss =\ $UNIMO$(batch)$
		\State$loss=v\_loss+ l\_loss+ cmcl\_loss$
		\State$loss.backward()$
		\EndFor
		\EndFunction
		
		\\
		\State \leftline{\color{gray}{\# \texttt{\scriptsize{build CMCL samples for each image-text pair}}}}
		\Function {$\mathbf{cmcl\_data\_loader}$}{}
		\State $samples = []$
		\State \leftline{\color{gray}{\# \texttt{\scriptsize{sample $a$ positive pairs from back-translation}}}}
		\State$pos\_pairs = sample\_pos\_pairs(pair,\ a)$
		\State \leftline{\color{gray}{\# \texttt{\scriptsize{sample $b$ negative pairs from text rewriting}}}}
		\State$neg\_pairs = sample\_neg\_pairs(pair,\ b)$
		\State \leftline{\color{gray}{\# \texttt{\scriptsize{sample $c$ sentences from text retrieval}}}}
		\State$pos\_imgs = sample\_pos\_imgs(pair,\ c)$
		\State \leftline{\color{gray}{\# \texttt{\scriptsize{sample $d$ images from image retrieval}}}}
		\State$pos\_texts = sample\_pos\_texts(pair,\ d)$
		\State $samples.extend(pair)$
		\State $samples.extend(pos\_pairs)$
		\State $samples.extend(neg\_pairs)$
		\State $samples.extend(pos\_imgs)$
		\State $samples.extend(pos\_texts)$
		\\ \ \ \ \ \ \ \Return $samples$
		\EndFunction
	}
	\end{algorithmic}  
\end{algorithm}

\section{Finetuning Settings}
\label{sec:finetune}

\paragraph{Task Definition and Details}
The multi-modal finetuning tasks include:
(1) \textbf{VQA} requires the model to answer natural language questions by selecting the correct answer from a multi-choice list based on an image. We conduct experiments on the widely-used VQA v2.0 dataset, which is built based on the COCO images. Similar to previous work, both training and validation sets are used for training for the results on both the test-std and test-dev splits.
(2) \textbf{Image Caption} requires the model to generate a natural language description of an image. We report our results on the Microsoft COCO Captions dataset. Following Karpathy's split, the dataset contains 113.2k/5k/5k images for train/val/test splits respectively. 
(3) \textbf{Visual Entailment (SNLI-VE)} is evaluated on the SLNI-VE dataset which was derived from Flickr30K images and Stanford Natural Language Inference (SNLI) dataset. The task is to determine the logical relationship (i.e., ``Entailment'', ``Neutral'' and ``Contradiction'') between a natural language statement and an image.
(4) \textbf{Image-Text Retrieval} is evaluated on the Flickr30k dataset, which contains two sub-tasks: image retrieval (Flickr30k-IR) and text retrieval (Flickr30k-TR), depending on which modality is used as the retrieved target. We report the top-K retrieval results on the test sets, including R@1, R@5 and R@10 (R denotes Recall).
The statistics of the datasets for the above multimodal-tasks are described in Table \ref{finetune-dataset}. 
The hyper-parameters for all the downstream tasks, including both the multi-modal tasks and single-modal tasks are shown in Table \ref{multimodal-params} and \ref{gen-params}.


\section{Visualization and Analysis}
\label{sec:visualization}

To intuitively show the effectiveness of the unified-modal learning on the corpus of images, texts and image-text pairs, we utilize 2-dimensional visualization of the embeddings by Principal component analysis (PCA). 
The nearest neighbors of the center word are shown in the embedding space.
UNIMO is compared with two ablation models described in Section 4.3.
The figure shows that the model ``UNIMO-w/o texts'' can find more visual relevant words than ``UNIMO-w/o image\&pairs'', which demonstrates the effectiveness of the visual learning on images.
However, UNIMO not only finds many visually relevant words, but also finds some semantic relevant background words.
For example, UNIMO finds ``lunch'' and ``airplanes'' for the center word ``hamburger'', which denotes people usually eat hamburger at lunch and often eat it while flying.
Also, for the second example, UNIMO finds relevant concepts ``meter'', ``steps'' and ``soccer'' for ``foot'', which enrich the concept and connect it with rich relevant information.

\begin{figure*}[t!]
	\centering
	\includegraphics[width=6.3in]{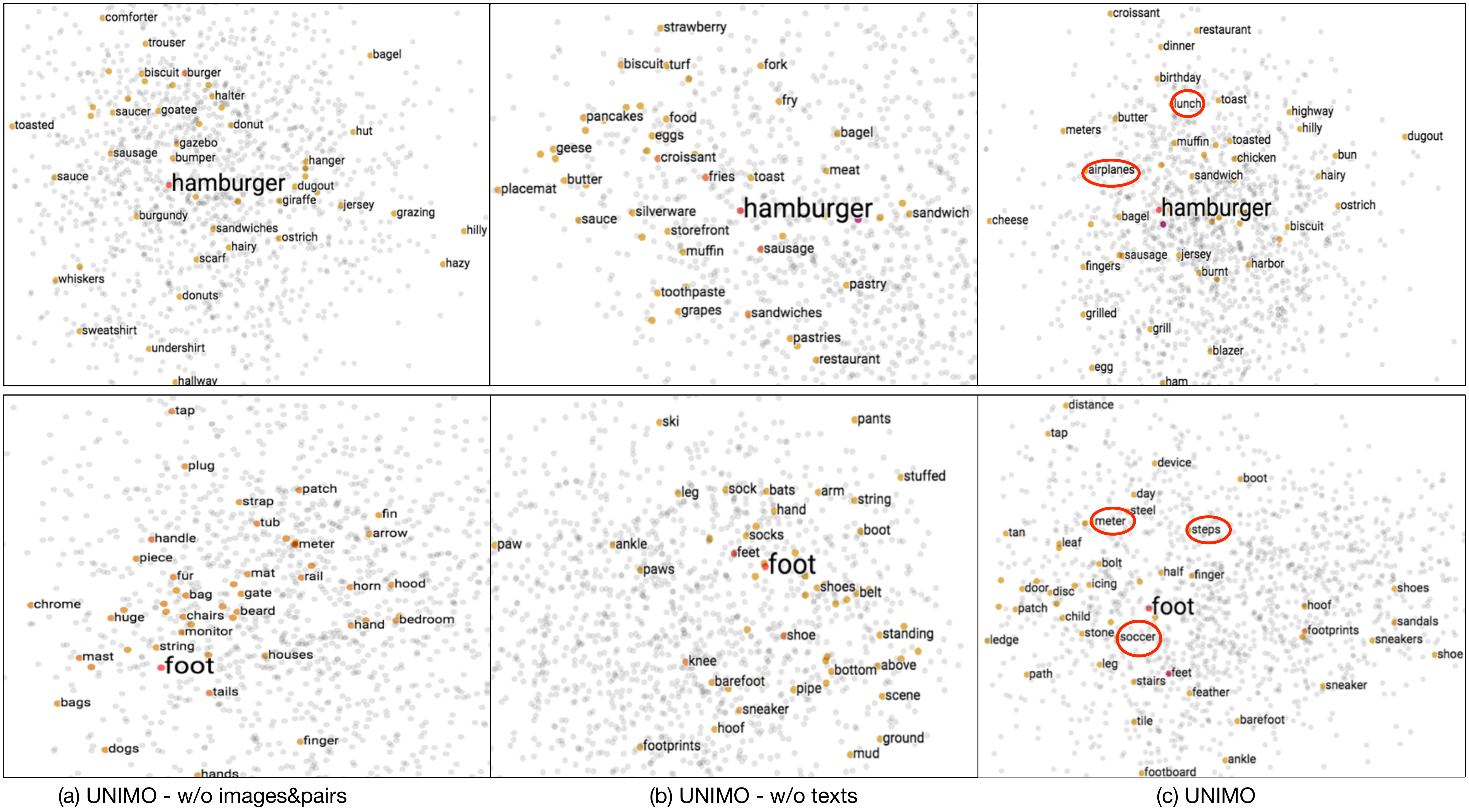}
	\caption{2-dimensional visualization by PCA.}
	\label{fig:visualization}
\end{figure*}

\begin{figure}[t!]
    \centering
    \includegraphics[width=3in]{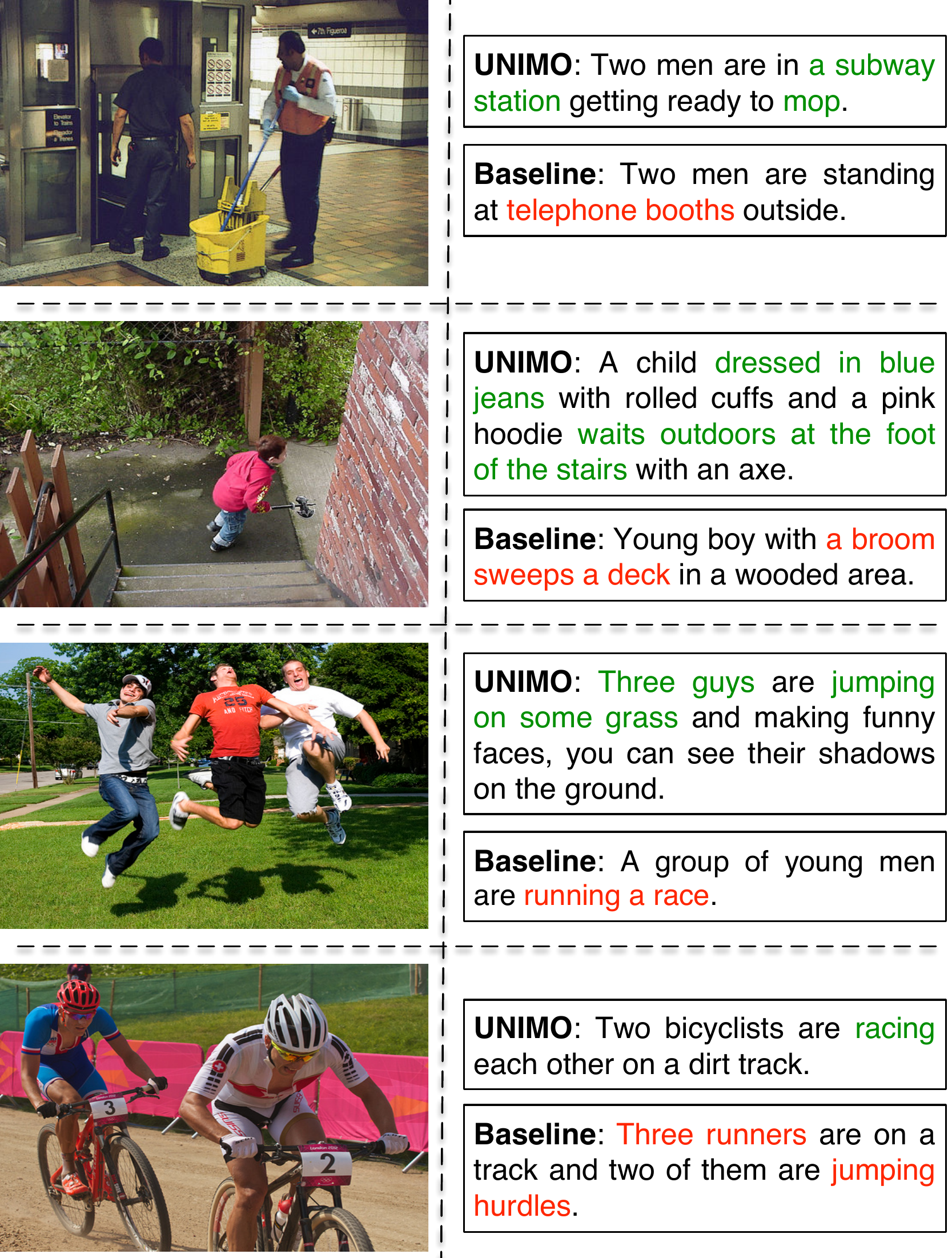}
    \caption{Text retrieval examples by R@1. The green color denotes accurate visual information while the red denotes wrong information.}
    \label{fig:text-retrieval}
\end{figure}

\begin{figure}[t!]
    \centering
    \includegraphics[width=3in]{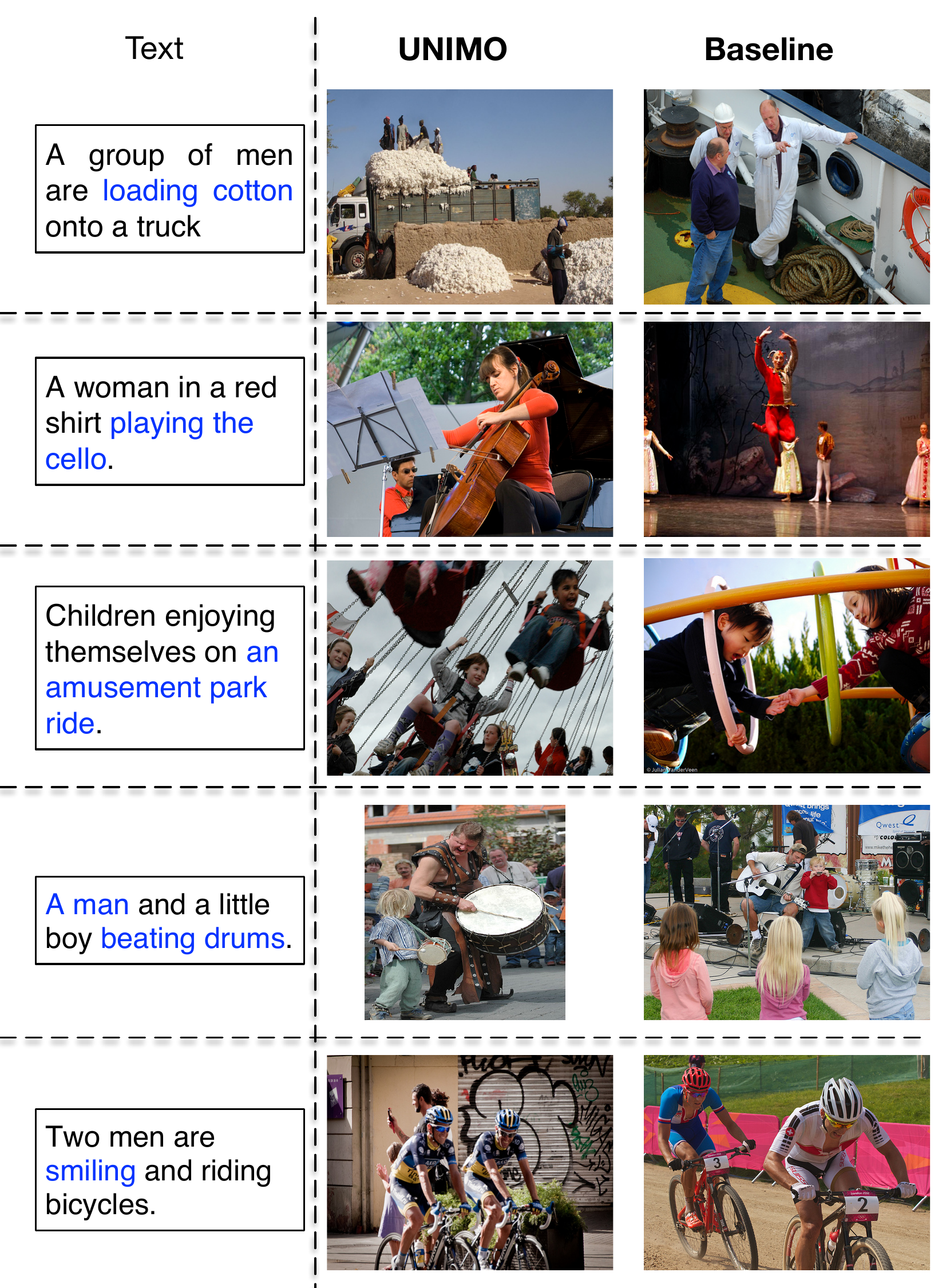}
    \caption{Image retrieval examples by R@1. The blue color denotes the important information that has been neglected by the baseline model, but is accurately recognized by UNIMO.}
    \label{fig:image-retrieval}
\end{figure}

To further intuitively show the advantages of the unified-modal learning with rich single-modal data, we compare UNIMO with the multimodal pre-training model ``w/o single modal'' (described in Section 4.2), on both the text retrieval and image retrieval tasks.
The examples of text retrieval results in Figure \ref{fig:text-retrieval} show that the retrieved captions by UNIMO describes the images more accurately by including different levels of information, including objects, attributes and relations in images.
The examples of the image retrieval results in Figure \ref{fig:image-retrieval} also show that the retrieved images better match the captions with more detail semantic alignments.

\end{document}